\documentclass[sigconf]{aamas} 

\usepackage{balance} % for balancing columns on the final page
\usepackage{bm}
\usepackage{algorithm}  
\usepackage{algorithmic}  
\usepackage{subfig}
\usepackage{url}
\usepackage{appendix}
\usepackage{amsmath}
\usepackage{diagbox}
\newtheorem{theorem}{\textbf{Theorem}}
\setcopyright{ifaamas}
\acmConference[AAMAS '21]{Proc.\@ of the 20th International Conference on Autonomous Agents and Multiagent Systems (AAMAS 2021)}{May 3--7, 2021}{Online}{U.~Endriss, A.~Now\'{e}, F.~Dignum, A.~Lomuscio (eds.)}
\copyrightyear{2021}
\acmYear{2021}
\acmDOI{}
\acmPrice{}
\acmISBN{}

\acmSubmissionID{32}

\title[AAMAS-2021 Formatting Instructions]{Modeling the Interaction between Agents in Cooperative  Multi-Agent Reinforcement Learning}

\author{Xiaoteng Ma$^*$}\thanks{$^*$Equal contribution.}
\affiliation{
  \institution{Tsinghua University}
  \city{Beijing, China}}
\email{ma-xt17@mails.tsinghua.edu.cn}

\author{Yiqin Yang$^*$}
\affiliation{
  \institution{Tsinghua University}
  \city{Beijing, China}}
\email{yangyiqi19@mails.tsinghua.edu.cn}

\author{Chenghao Li$^*$}
\affiliation{
	\institution{Tsinghua University}
	\city{Beijing, China}}
\email{lich18@mails.tsinghua.edu.cn}

\author{Yiwen Lu}
\affiliation{
	\institution{Tsinghua University}
	\city{Beijing, China}}
\email{luyw20@mails.tsinghua.edu.cn}

\author{Qianchuan Zhao}
\affiliation{
  \institution{Tsinghua University}
  \city{Beijing, China}}
\email{zhaoqc@tsinghua.edu.cn}

\author{Yang Jun}
\affiliation{
  \institution{Tsinghua University}
  \city{Beijing, China}}
\email{yangjun603@tsinghua.edu.cn}

\begin{abstract}
Value-based methods of multi-agent reinforcement learning (MARL), especially the value decomposition methods, have been  demonstrated on a range of challenging cooperative tasks.
However, current methods pay little attention to the interaction between agents, which is essential to teamwork in games or real life. 
This limits the efficiency of value-based MARL algorithms in the two aspects: collaborative exploration and value function estimation. 
In this paper, we propose a novel cooperative MARL algorithm named as interactive actor-critic~(IAC), which models the interaction of agents from the perspectives of policy and value function. 
On the policy side, a multi-agent joint stochastic policy is introduced by adopting a collaborative exploration module, which is trained by maximizing the entropy-regularized expected return. 
On the value side, we use the shared attention mechanism to estimate the value function of each agent, which takes the impact of the teammates into consideration.
At the implementation level, we extend the value decomposition methods to continuous control tasks and evaluate IAC on benchmark tasks including classic control and multi-agent particle environments.
Experimental results indicate that our method outperforms the state-of-the-art approaches and achieves better performance in terms of cooperation.
\end{abstract}


\newcommand{\BibTeX}{\rm B\kern-.05em{\sc i\kern-.025em b}\kern-.08em\TeX}

\begin{document}

%%% The following commands remove the headers in your paper. For final 
%%% papers, these will be inserted during the pagination process.

\pagestyle{fancy}
\fancyhead{}

%%% The next command prints the information defined in the preamble.

\maketitle 

%%%%%%%%%%%%%%%%%%%%%%%%%%%%%%%%%%%%%%%%%%%%%%%%%%%%%%%%%%%%%%%%%%%%%%%%

\section{Introduction}\label{introduction}
In recent years, cooperative multi-agent reinforcement learning is an active research area~\cite{hernandez2019survey,zhang2018fully,cassano2020multi, bertsekas2020multiagent} and has made exciting progress in many domains, including traffic signal network optimization~\cite{zhang2020integrating,sykora2020multi,chu2019multi} and network packet routing~\cite{mao2019modelling}. The full spectrum of cooperative MARL algorithms is a hybrid of 
\textit{centralized training and decentralized execution}~(CTDE)~\cite{oliehoek2008optimal}. In the CTDE framework, value-based methods receive the most attention and they can be divided into two categories: centralized value function methods and value decomposition methods~\cite{yang2020q}. 

The centralized value function methods allow agents to learn approximate models of other agents online and further effectively use estimated models in their policy learning procedure. For example, MADDPG employs the deterministic policies of other agents to learn the centralized value function~\cite{lowe2017multi}. Counterfactual multi-agent policy gradient~(COMA) adopts a counterfactual baseline to marginalize out a single agent’s action while keeping the other agents’ actions fixed ~\cite{foerster2018counterfactual}. MADDPG and COMA introduce respectively the deterministic policy gradient and the policy gradient methods into MARL. 
Unfortunately, the suboptimality of one agent’s policy can propagate through the centralized value function and negatively affect policy learning of other agents, which is named \textit{centralized-decentralized mismatch}~(CDM)~\cite{wang2020off}. 
%This problem causes a discrepancy in performance between multi-agent policy gradient and value function decomposition methods. 
To alleviate the problem, Multi Actor Attention Critic~(MAAC) adopts a shared attention mechanism, a successful approach applied in various domains~\cite{vaswani2017attention, ba2014multiple, bahdanau2015neural, oh2016control}, to select relevant value function information of agents to calculate the centralized value function~\cite{iqbal2019actor}. 
However, MAAC has a limited performance improvement compared with MADDPG and cannot get rid of the CDM problem. 

Different from centralized value function methods, current value decomposition methods adopt structural constraints to meet that local maxima on the per-agent action-values amount to the global maximum on the joint action-value, which is named Individual-Global-Max~(IGM) assumption.
Value-Decomposition Networks~(VDN) proposes a learned additive value-decomposition approach over individual agents, which aims to learn an optimal linear value decomposition from the team reward signal~\cite{sunehag2017value}. 
Another example is QMIX, which enforces a monotonic structure between the joint action-value function and the per-agent action-value functions.
This structure allows tractable maximization of the joint action-value function in off-policy learning and guarantees consistency between the centralized and decentralized policies ~\cite{rashid2018qmix}. 
Moreover, QTRAN proposes a new factorization method, which can transform the original joint action-value function into an easily factorizable one, with the same optimal actions~\cite{son2019qtran}. 
These value decomposition methods can effectively estimate the joint action-value function by calculating individual action-value functions, and achieve better performance than centralized value function methods.

%However, existing methods ignore one key challenge for designing MARL algorithms: the interaction between agents. 
Compared with single-agent RL, the growth of the agents' number in MARL causes the exponential growth of the state-action space. 
This phenomenon seriously affects the estimation accuracy of the value function, which is conspicuous for centralized value function methods.
The variance of multi-agent policy gradient would grow exponentially as the number of agents increases~\cite{wang2020off}.
% This causes a discrepancy in performance between multi-agent policy gradient and value function decomposition methods.
% The centralized value function methods steady-state distribution in which every state has a strictly positive probability of visitation~\cite{sayed2014adaptation}.
%Value function decomposition methods avoid the curse of dimensionality in action space by assuming that the value function of each agent is independent, while the IGM assumption constrains the representation ability of global value function~\cite{iqbal2020ai}. 
Value decomposition methods avoid the curse of dimensionality in estimating joint action-value function by assuming that the value function of each agent is independent, while the IGM assumption limits the representation power of joint action-value function~\cite{iqbal2020ai}. 
For example, in the traffic system at the crossroads, if the running car only maximizes self expected reward, the transportation system may fall into chaos~\cite{al2019cooperative}. 
Therefore, the optimization objective of agents in cooperative tasks should contain cumulative self expected reward and cooperative expected reward. 
We name the newly defined expected objective as \textit{mutual value function information}~(MVFI). 

From the policy view, another problem is the collaborative exploration, which is neglected in current MARL algorithms. 
When directly applying the existing MARL algorithms to complex games with a large number of agents, agents may fail to learn good strategies and end up with little interaction with other agents even when collaboration is significantly beneficial~\cite{long2019evolutionary}. 
%Although some works in single-agent exploration achieve success~\cite{jin2018q, sayed2014adaptation, osband2016lower, agrawal2017optimistic,pathak2017curiosity, ostrovski2017count}, 
Back to single-agent RL, maximum entropy policy achieves conspicuous success in exploration by acquiring diverse behaviors.
Some works have revealed that the objective with entropy regularization enjoys the smoother optimization landscape and the faster convergence rate, which builds connections between RL with probabilistic graphical models and convex optimization~\cite{ahmed2019understanding}.
Therefore, directly introducing the maximum entropy policy into MARL is a natural idea, which is already explored~\cite{haarnoja2017reinforcement}.
However, this method just aims to impose the maximum entropy policy on the individual policies rather than the joint stochastic policy. 
Similar work has done to generalize maximum entropy policy into the stochastic games~\cite{grau2018balancing}, such as the two-player soft q-learning~(TPSQL).
Although this work shows that the stochastic game with soft Q-learning would exhibit an optimal state value, TPSQL is restricted to the two-player zero-sum stochastic games. 
For this reason, designing an effective multi-agent maximum entropy algorithm remains an open research problem.

\textbf{Contributions.} In this paper, we propose an novel off-policy cooperative MARL algorithm, named IAC. The full architecture is discussed in detail in Section~\ref{Collaborative Exploration MARL}.
Our method has three following key technical and experimental contributions. We analyze the necessity and challenge of modeling the interaction between agents.
To deal with the problems, we propose a multi-agent joint stochastic policy by adopting a collaborative exploration module on the policy side, which is trained by maximizing the entropy-regularized expected return.
On the value side, we propose a new optimization objective named mutual value function information, which is estimated by the shared attention mechanism.
As for the implementation issue, we extend the value decomposition methods to continuous control tasks and evaluate our algorithm on benchmark tasks including classic control and multi-agent cooperative tasks.
Experimental results are shown in Section~\ref{Experiment result}, which demonstrates that IAC not only outperforms the state-of-the-art approaches but also achieves better performance in terms of cooperation, robustness, and scalability. 
Moreover, the positive performances of multi-agent joint stochastic policy and MVFI are confirmed respectively in experiments.
Our work appears to be the first study of modeling the interaction between agents under CTDE framework in cooperative MARL.

\section{Background}
In this section, we give the introduction to the problem formulation of multi-agent cooperation and the basic building blocks for our approach.

\subsection{Notation}
We consider Dec-POMDP~\cite{oliehoek2016concise} as a standard model consisting of a tuple $\mathcal{G}= \langle \mathcal{S},\mathcal{U},\mathcal{P},\mathcal{R},\mathcal{Z},\mathcal{O},\gamma  \rangle $ for cooperative multi-agent tasks. 
Within $\mathcal{G}$, $s\in \mathcal{S}$ denotes the global state of the environment. 
Each agent $i\in N:={1,...,n}$ chooses an action $u_i\in \mathcal{U}$ at each time, forming a joint action $\textbf{u}\in \mathcal{U}^n$. 
Let $\mathcal{P}(s'\mid s,\textbf{u}):\mathcal{S}\times \mathcal{U}^n\times \mathcal{S}\to [0,1]$ denotes the state transition function.
The reward function $\mathcal{R}(s,\textbf{u}):\mathcal{S}\times \mathcal{U}^n\to \mathbb{R}$ is shared among all agents and $\gamma \in [0,1)$ is the discount factor. 

In a partially observable scenario, each agent has individual observations $z\in \mathcal{Z}$ according to the observation function $\mathcal{Z}(s,u):\mathcal{S}\times \mathcal{U} \to \mathcal{O}$, which conditions on a stochastic policy $\pi_i(u_i \mid o_i)$ parameterized by $\theta_i$: $\mathcal{O}\times \mathcal{U} \to [0,1]$. 
Each agent has its observation history $\tau_i\in \mathcal{T}:=(\mathcal{O}\times \mathcal{U})^t$.
The joint action-value function is defined as
\begin{equation}
Q^{\bm{\pi}}_{\rm{tot}}(s,\bm{u}) = \mathbb{E}_{s_t, \bm{u}_t\sim \bm{\tau}_{\bm{\pi}}} \left[\sum_{t=0}^{T}\gamma^t r_{t} \mid s_t=s,\bm{u}_t=\bm{u}\right], 
\end{equation}
where $\bm{\pi}$ is the joint policy with parameters $\theta= \langle \theta_1,\dots,\theta_n  \rangle $.

\subsection{Value Decomposition Methods}\label{IGM-section}
Value decomposition methods are widely used in value-based MARL algorithms~\cite{rashid2020weighted, iqbal2020ai}. These methods estimate the joint action-value function by learning individual action-value functions. 
For a joint action-value function $Q^{\bm{\pi}}_{\rm{tot}}(\bm{o},\textbf{u})$, if the following holds
\begin{align}\label{IGM}
\mathop{ \arg\max}_{\textbf{u}}Q^{\bm{\pi}}_{\rm {tot}}(\bm{o},\textbf{u}) & =  
\left(                 %左括号
\begin{array}{c}   %该矩阵一共3列，每一列都居中放置
\mathop{ \arg\max}_{u_1}Q_1(o_1,u_1)\\  %第一行元素
\vdots\\  %第二行元素
\mathop{ \arg\max}_{u_n}Q_n(o_n,u_n)\\
\end{array}
\right),                 %右括号
\end{align}
$Q_i$ satisfies $\textbf{IGM Condition}$ and the $Q^{\bm{\pi}}_{\rm{tot}}(\bm{o},\textbf{u})$ can be factorized by $Q_i(o_i, u_i)$. 
Equation~\ref{IGM} demonstrates that local maxima on the individual action-values amount to the global maximum on the joint action-value.

Building on purely independent DQN-style agents, VDN~\cite{sunehag2017value} assumes that the joint action-value function can be additively decomposed into action-value functions across agents.
Differently, QMIX~\cite{rashid2018qmix} applies a structural constraint between $Q^{\bm{\pi}}_{\rm{tot}}(\bm{o},\textbf{u})$ and $Q_i(o_i, u_i)$ to meet the monotonic assumption
\begin{eqnarray}\label{QMIX}
\frac{\partial Q^{\bm{\pi}}_{\rm{tot}}(\bm{o},\textbf{u})}{\partial Q_i(o_i, u_i)} \ge 0, {\forall}i\in N.
\end{eqnarray}
The monotonic condition is guaranteed by a mixing network with nonnegative weights, which in turn guarantees Equation~\ref{IGM}.
Massive challenging MARL tasks, such as StarCraft~II, demonstrate the monotonic decomposition form is simple but effective as it can be performed in $\mathcal{O}(n\left|\mathcal{U}\right|)$ time as opposed to $\mathcal{O}(|\mathcal{U}|^n)$.
%However, structure constraints of value decomposition methods are criticized due to the limited representation capacity of the joint action-value function. 
%Many works explore to deal with the problem.
%For example, Multi-agent determinantal q-learning~(Q-DPP) applies the determinantal point process to acquire diverse behavioral models~\cite{yang2020multi}. 
%Q-DPP allows a natural factorization of the joint Q-functions with no need for a priori structural constraints on the value function or special network architectures. 
%Unfortunately, Q-DPP needs to record all possible observation-action pairs of agents, resulting in infeasible implementation.

\subsection{Maximum Entropy RL}
Maximum Entropy Learning is an active research area in recent years and it has been applied in many aspects~\cite{ziebart2008maximum,todorov2008general}. 
Different with the standard reinforcement learning, which maximizes the expected sum of rewards $\sum_{t=0}^{T}\mathbb{E}_{(o_t,u_t)\sim \tau_{\pi}}[r(o_t,u_t)]$, the maximum entropy policy favors stochastic policies by augmenting the objective with the expected entropy of the policy
\begin{align}
J(\pi) &= \sum_{t=0}^{T}\mathbb{E}_{(o_t,u_t)\sim \tau_{\pi}}\left[r(o_t,u_t)+\alpha \mathcal{H}(\pi(\cdot \mid o_t))\right],
\end{align}
where $\mathcal{H}(\pi(\cdot \mid o_t))$ denotes the entropy of the policy, which implies the randomness of the policy. 
The temperature parameter $\alpha$ determines the relative importance of the entropy term against the reward. 
Maximum entropy policy calculates the soft Q-value iteratively with the entropy-regularized expected return objective as follows:
\begin{align}\label{Q_soft}
Q_{\rm{soft}}(o_t,u_t) &= r(o_t, u_t) + \gamma \mathbb{E}_{o_{t+1}\sim p}\left[V_{\rm{soft}}(o_{t+1})\right], \\
V_{\rm{soft}}(o_t) &= \mathbb{E}_{u_t\sim \pi}\left[Q_{\rm{soft}}(o_t, u_t)-\alpha\log \pi(u_t \mid o_t)\right].
\end{align}
According to the learned Q-function in Equation~\ref{Q_soft}, maximum entropy policy is directly updated for each state by  minimizing the following objective
\begin{align*}
J(\pi) &= \mathbb{E}_{o_t\sim \tau_{\pi}}\left[ D_{KL}\left(\pi(\cdot \mid o_t)\parallel \frac{\exp\left(\frac{1}{\alpha}Q_{\rm{soft}}\left(o_t,\cdot\right)\right)}{Z_{\rm{soft}}(o_t)}\right)\right] \notag \\
&= \mathbb{E}_{o_t\sim \tau_{\pi},\epsilon_t \sim \mathcal{N}}\left[\log\pi\left(f(\epsilon_t;o_t)\mid o_t\right)-\frac{1}{\alpha} Q_{\rm{soft}}\left(o_t,f(\epsilon_t;o_t)\right)\right],
\end{align*}
where $Z_{\rm{soft}}(o_t)$ denotes the partition function and $\epsilon_t$ represents the noise vector sampled in Gaussian distribution $\mathcal{N}$. The policy is reparameterized~\cite{1992Simple} by a neural network transformation $u_t=f(\epsilon_t;o_t)$. The above iteration process leads to an improved policy in terms of the soft value function~\cite{haarnoja2018soft}. 

%Maximum entropy learning is based on the Energy-Based Model~(EBM), which formulates the equilibrium probabilities via the Boltzmann distribution~\cite{sallans2004reinforcement, levine2014learning}
%One successful application of EBM in single-agent RL is adopting the Boltzmann distributions to approximate policies~\cite{levine2014learning}.
%Differently, the maximum entropy policy maximizes both the expected return and the expected entropy. 

%The maximum entropy distribution could be used to guide policy learning towards high-reward regions.
%Soft Q-Learning~(SQL)~\cite{haarnoja2017reinforcement} and Soft Actor-Critic~(SAC)~\cite{haarnoja2018soft} are two typical algorithms, which achieves state-of-art performance on a range of single-agent continuous control benchmark tasks. 
%Similar to SQL and SAC algorithm, IAC maximizes the entropy-regularized expected return by adopting a multi-agent joint stochastic policy. 
%Also, IAC is trained under the CTDE framework and enjoys the superiority of maximum entropy theory.

\subsection{Collaborative Exploration in MARL}
Exploration is necessary due to the long horizons and limited or delayed reward signals in complex tasks.
In general, current exploration methods in single-agent RL mainly include heuristics methods~\cite{houthooft2016vime}, such as $\epsilon$-greedy and count-based methods~\cite{bellemare2016unifying, tang2017exploration}. 
However, most algorithms for statistically efficient RL are not computationally tractable in complex environments~\cite{osband2016deep}.
To deal with this problem, MAVEN introduces a latent space for hierarchical control, which would be fixed over an entire episode. 
Each joint action-value function can be thought of as a mode of joint exploratory behavior~\cite{maven}. 
Unfortunately, it is extremely hard to guarantee the validness and diversity of the learned latent space.

\section{Analysis of Interaction in MARL}\label{Analysis}
Current MARL algorithms pay little attention to the interaction between agents. 
For example, individual exploration and individual action-value function estimation in QMIX.
In this section, we analyze the case that the policy learned by QMIX cannot represent the true optimal action-value function as well as cannot be the optimal policy.
A simple example is given by the payoff matrix of the two-player three-action matrix game, which is shown in Table~\ref{Payoff}. 
\begin{table}[H]
%\begin{minipage}{0.43\linewidth}
    \centering
    \begin{tabular}{|c||c|c|c|}
    \hline
    \diagbox[width=2.5em,trim=l]{$u_2$}{$u_1$} & \textbf{A} & B & C \\ \hline
    \hline
    \textbf{A} & \textbf{8} & -12 & -12 \\
    \hline
    B & -12 & 0 & 0 \\
    \hline
    C & -12 & 0 & 0 \\
    \hline
    \end{tabular}
    \caption{Payoff of matrix game}
    \label{Payoff}
 %   \end{minipage}
    
    %\begin{minipage}{0.44\linewidth}  
    \centering
    \begin{tabular}{|c||c|c|c|}
    \hline
    \diagbox[width=3.0em,trim=l]{$Q_2$}{$Q_1$} & -8.9(A) & -0.6(B) & \textbf{-0.5(C)} \\ \hline
    \hline
    -8.0(A) & -8.06 & -8.05 & -8.05 \\
    \hline
    0.1(B) & -8.05 & -0.01 & -0.01 \\
    \hline
    \textbf{0.2(C)} & -8.05 & -0.02 & \textbf{-0.01} \\
    \hline
    \end{tabular}
    \caption{Individual and joint action-value of QMIX}
    \label{QMIX}
%\end{minipage}
\end{table}

According to the learned values in Table~\ref{QMIX}, the Boltzmann policy of QMIX is shown in Table~\ref{Policy QMIX} according to 
\begin{equation}
    P(u_{i,j})=\frac{\exp(Q_{i,j}/\alpha)}{\sum_{i,j}\exp(Q_{i,j}/\alpha)}.
\end{equation}
However, experimental result indicates that QMIX learns a suboptimal policy under uniform visitation.
To solve this problem, several methods choose to communicate self-state information with other agents~\cite{zhang2019efficient, das2019tarmac}. But this does not work in the grid task as there is no state-information. 
Meanwhile, we find out that if each agent knows the actions of other agents in advance and use them as part of the state, we can accurately get the optimal solution, which is shown in Table~\ref{Policy QMIXs}. 
For the convenience of expression, we name this method as QMIXs. 
For details of it, please refer to the Appendix~\ref{QMIXs}.
Unfortunately, interacting action-information violates the Dec-POMDP framework where all agents choose actions simultaneously.  
For this reason, we propose an novel MARL algorithm modeling the interaction between agents under the Dec-POMDP framework in Section~\ref{Collaborative Exploration MARL}.
\begin{table}[H]
\begin{minipage}{0.47\linewidth}
    \centering
    \begin{tabular}{|c||c|c|c|}
    \hline
    \diagbox[width=2.5em,trim=l]{$u_2$}{$u_1$} & A & B & C \\ \hline
    \hline
    A & 0.01 & 0.02 & 0.02 \\
    \hline
    B & 0.02 & 0.23 & 0.23 \\
    \hline
    C & 0.01 & 0.23 & 0.23 \\
    \hline
    \end{tabular}
    \caption{Policy of QMIX}
    \label{Policy QMIX}
    \end{minipage}
    \begin{minipage}{0.47\linewidth} 
    \centering
    \begin{tabular}{|c||c|c|c|}
    \hline
    \diagbox[width=2.5em,trim=l]{$u_2$}{$u_1$} & \textbf{A} & B & C \\ \hline
    \hline
     \textbf{A} & \textbf{0.76} & 0.01 & 0.01\\
    \hline
    B & 0.01 & 0.05 & 0.05 \\
    \hline
    C & 0.01 & 0.05 & 0.05\\
    \hline
    \end{tabular}
    \caption{Policy of QMIXs}
    \label{Policy QMIXs}
\end{minipage}
\end{table}

\section{An Interactive Approach for MARL}\label{Collaborative Exploration MARL}
\begin{figure*}[t]
	\centering
	\includegraphics[width=5.6in]{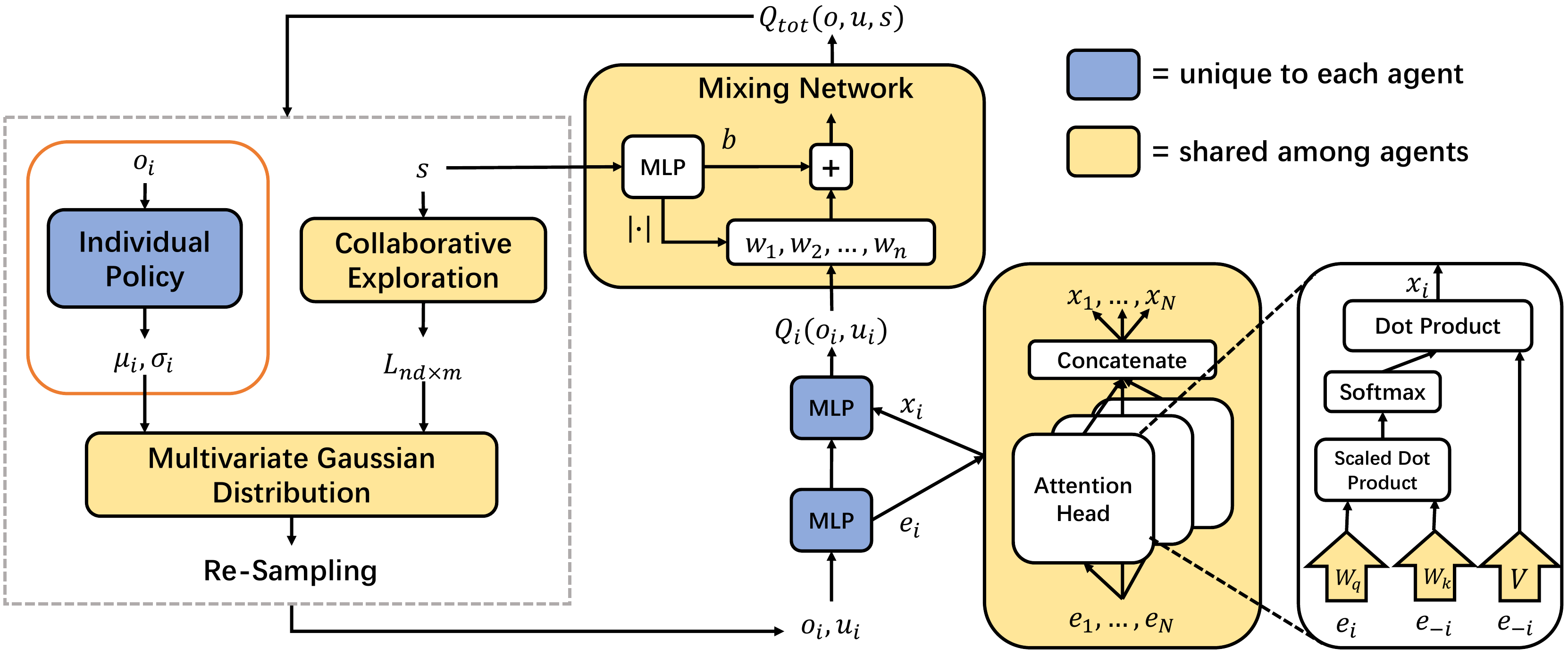}
	\caption{The structure of IAC. The leftmost box denotes the collaborative exploration module. Attention-QMIX Structure is shown on the right-hand side.
	During the centralized training, all the boxes are activated with shared information. During the decentralized execution, agents observe local information and only the orange boxes are activated.
    }
	\label{SAQ}
	%\Description{Logo showing the words "AAMAS 2021, London, UK", 
	%with the 0 resembling the London Eye and the 1 resembling Big Ben.}
\end{figure*}

In this section, we analyze the necessity for modeling the interaction between agents and the key challenges of the maximum entropy MARL.
Based on these challenges, we introduce the novel cooperative MARL algorithm, \textbf{Interactive Actor-Critic}, which is shown in Figure~\ref{SAQ}.
The complete algorithm is illustrated in Algorithm~1. 

\subsection{Collaborative Exploration Module}\label{CE Module}
Individual exploration in single-agent RL, such as increasing per-agent exploration rate $\epsilon$, can help exploration. 
However, individual exploration may lead to provably poor exploration and suboptimality in MARL~\cite{maven}, which is illustrated by the simple example in Section~\ref{Analysis}. 
As the superior exploration ability of the maximum entropy policy discussed in the Section~\ref{introduction}, a natural idea is extending the maximum entropy policy to MARL algorithms. 
The modified objective with the expected entropy of the joint stochastic policy is defined as
\begin{eqnarray}\label{Ent}
\bm{\pi}^* = \mathop{ \arg\max}_{\bm{\pi}}\sum_{t=0}^{T}\mathbb{E}_{(\bm{o}_t,\bm{u}_t)\sim \bm{\tau}_{\bm{\pi}}}\left[r(\bm{o}_t,\bm{u}_t)+\alpha \mathcal{H}(\bm{\pi}(\cdot \mid s_t))\right].
\end{eqnarray}
With the joint action-value function, the joint stochastic policy is updated as follows:
\begin{eqnarray}\label{pi_new}
\bm{\pi}_{\rm{new}} = \mathop{ \arg\min}_{\bm{\pi}'} D_{KL}\left(\bm{\pi}'(\cdot \mid \bm{o}_t) \parallel \frac{\exp\left(\frac{1}{\alpha}Q^{\bm{\pi}_{\rm{old}}}_{\rm{tot}}(\bm{o}_t,\cdot)\right)}{Z^{\bm{\pi}_{\rm{old}}}_{\rm{tot}}\left(\bm{o}_t\right)}\right).
\end{eqnarray}
The above iteration is the standard multi-agent maximum entropy policy, and it mainly follows along the standard Soft Q-Learning theorem~\cite{haarnoja2017reinforcement}. 
Following the proof of soft policy iteration in SAC~\cite{haarnoja2018soft}, this standard maximum entropy policy in MARL inherits the convergence property for the joint stochastic policy iteration, which is formally stated in Theorem~\ref{theorem}.

\begin{theorem}\label{theorem}
Let $\bm{\pi}_{\rm{old}}$ and $\bm{\pi}_{\rm{new}}$ denote the joint stochastic policy, which is the optimizer of the minimization problem in Equation~\ref{pi_new}.
The joint action-value function has a joint stochastic policy improvement $Q_{\rm{tot}}^{\bm{\pi}_{\rm{new}}}(\bm{o}_t,\bm{u}_t) \ge Q_{\rm{tot}}^{\bm{\pi}_{\rm{old}}}(\bm{o}_t,\bm{u}_t)$ for all $(\bm{o}_t, \bm{u}_t) \in \mathcal{O}\times \mathcal{U}$ with  $|\mathcal{U}|<\infty$. 
The joint stochastic policy $\bm{\pi}$ could converge to $\bm{\pi}^*$ such that $Q_{\rm{tot}}^{\bm{\pi}^*}(\bm{o}_t,\bm{u}_t) \ge Q_{\rm{tot}}^{\bm{\pi}}(\bm{o}_t,\bm{u}_t)$ for all $\bm{\pi}\in \Pi$, where $\Pi$ denotes the set of policies.
\end{theorem}

The standard maximum entropy MARL method learns the joint stochastic policy $\bm{\pi}(\bm{u}_t \mid \bm{o}_t)$ while the individual policy $\pi_i(u_i \mid o_i)$ is unavailable, which violates the CTDE framework. To deal with the issue, we attempt to use the multivariate Gaussian distribution $\mathcal{N}_{M}$ to model the interaction between individual policies because the behavioral strategies reflect agents' cooperation relationship~\cite{liu2019multi}.
Let $d$ denote the dimension of the action space for each agent and $\Sigma$ denote the covariance matrix of the multivariate Gaussian distribution. The probability density function of the distribution $\mathcal{N}_{M}$ is calculated as follows:
\begin{eqnarray}\label{f_M}
f(x)=\frac{1}{\left(\sqrt{2\pi}\right)^{nd}|\Sigma|^{\frac{1}{2}}}\exp\left\{{-\frac{(x-\mu_x)^T(\Sigma)^{-1}(x-\mu_x)}{2}}\right\},
\end{eqnarray}
where $x$ is a random variable that represents actions sampled from the multivariate Gaussian distribution. 
$\mu_x$ indicates the mean of $x$, which represents the output of individual policies. 
When the individual policies are independent, the off-diagonal elements of the covariance matrix $\Sigma$ are $\bm{0}$.
The joint action-value function can be simplified to
\begin{eqnarray}
Q_{\rm{tot}} = \sum_{t=0}^{T}\mathbb{E}_{(\bm{o}_t,\bm{u}_t)\sim \bm{\tau}_{\bm{\pi}}}\left[r(\bm{o}_t,\bm{u}_t)-\alpha \sum_{i=1}^{n}\log\pi_i(\cdot \mid o_t^i)\right].
\end{eqnarray}

Otherwise, the covariance matrix $\Sigma$ is a real symmetric matrix, which reflects the relationship between behavioral strategies of agents. 
When the agent number increases largely, it is hard to approximate the value of $\Sigma$ due to the exponential growth of interactions and dimensionality. 
Therefore, we replace $\mathcal{N}_{M}$ with the \textit{low rank multivariate Gaussian distribution} $\mathcal{N}_{LM}$ and decompose the covariance matrix $\Sigma_{nd\times nd}$ as follows:
\begin{eqnarray} \label{simga_decompose}
\Sigma_{nd\times nd} = L_{nd\times m} L_{nd\times m}^T + D_{nd\times nd},
\end{eqnarray}
where $n$ is the number of agents and $L_{nd\times m}$ denotes the covariance factor. 
$D_{nd\times nd}$ denotes the covariance diagonal matrix. 
$L_{nd\times m}$ is named the \textit{collaborative exploration matrix}, which is approximated with a shared neural network
\begin{eqnarray}
L_{nd\times m} = f_{\psi}(s_t;\psi),
\end{eqnarray}          
where $f_{\psi}(s)$ is parameterized by $\psi$ and the input of the shared neural network is global state information $s$.
The joint stochastic policy is reparameterized as follows
\begin{align}\label{u_t}
\bm{u}_t &=  p_{\theta,\psi}(\bm{\epsilon}_N, \bm{\epsilon}_M; \bm{o}_t,s_t) \notag \\ \notag
&= f_{\theta}(\bm{\epsilon}_N;\bm{o}_t) + f_{\psi}(\bm{\epsilon}_M; s_t)   \\
&= \bm{\mu}_t + \bm{\sigma}_t\bm{\epsilon}_N + L_{nd\times m}\bm{\epsilon}_M,
\end{align}
where $p_{\theta,\psi}$ denotes the joint stochastic policy network composed of the individual policy network $f_{\theta}$ the collaborative exploration network $f_{\psi}$.
$\bm{u}=[\mu_1,\dots,\mu_n], \bm{\sigma}=[\sigma_1,\dots,\sigma_n]$ are the vectors, where $\mu_i$, $\sigma_i$ are the outputs of the individual policies $\pi_i(o_i;\theta_i)$.
$\bm{\epsilon}_N$ denotes the individual Gaussian noise and $\bm{\epsilon}_M$ denotes the collaborative Gaussian noise.

While executing, we set the individual Gaussian noise $\bm{\epsilon}_N$ and the collaborative Gaussian noise to $\bm{0}$. 
Each agent only applies their individual information to make decisions. 
The collaborative exploration policy only works in the training phase, which meets the requirement of the CTDE framework. 
Sine the joint stochastic action $\bm{u}$ is resampled from the low rank multivariate Gaussian distribution $\mathcal{N}_{LM}$, the entropy of the joint stochastic policy $\mathcal{H}(\bm{\pi}(\cdot \mid s_t))$ can be calculated according to the Equation~\ref{f_M}. 
The policy parameters are learned by minimizing the following equation
\begin{align}\label{collaborative exploration}
& J_{\bm{\pi}}(\theta,\psi) = \mathbb{E}_{ \bm{\tau}\sim{\bm{\pi}}} \left[D_{KL}\left(\bm{\pi}_{\theta,\psi}(\cdot \mid \bm{o}_t,s_t)\parallel \frac{\exp\left(\frac{1}{\alpha} Q^{\bm{\pi}_{\rm{old}}}(\bm{o}_t,\cdot)\right)}{Z^{\bm{\pi}_{\rm{old}}}(\bm{o}_t)}\right)\right] \notag \\
& = \mathbb{E}_{(s_t, \bm{o}_t)\sim \bm{\tau}_{\bm{\pi}}, \bm{\epsilon}_N,\bm{\epsilon}_M \sim \mathcal{N}} [\log\bm{\pi}_{\theta,\psi}(p_{\theta,\psi}(\bm{\epsilon}_N,\bm{\epsilon}_M;\bm{o}_t,s_t) \mid \bm{o}_t,s_t)- \notag \\
& \qquad \quad \frac{1}{\alpha} Q_{\rm{tot}}(\bm{o}_t,p_{\theta,\psi}(\bm{\epsilon}_N,\bm{\epsilon}_M;\bm{o}_t,s_t))].
\end{align}
%The gradient formulation of Equation~\ref{collaborative exploration} is 
%\begin{align}\label{Policy}
%& \nabla_{\theta_i}J_{\bm{\pi}}(\theta,\psi)=\nabla_{\theta_i}\log\bm{\pi}_{\theta,\psi}(\bm{u}_t|\bm{o}_t,s_t)- \notag \\
%&\qquad \quad \frac{1}{\alpha} \nabla_{u_i}Q_{\rm{tot}}(\bm{o}_t,u_i,u_{-i})\mid_{u_i=\pi_{\theta_i,\psi}(o_t^i,s_t)}\nabla_{\theta_i}\pi_{\theta_i,\psi}(o_t^i,s_t), \\
%& \nabla_{\psi}J(\bm{\pi})(\theta,\psi)=\nabla_{\psi}\log\bm{\pi}_{\theta,\psi}(\bm{u}_t \mid \bm{o}_t,s_t)- \notag \\
%&\qquad \quad \frac{1}{\alpha} \nabla_{\bm{u}_t}Q_{\rm{tot}}(\bm{o}_t,\bm{u}_t)\mid_{\bm{u}_t=\bm{\pi}_{\theta,\psi}(\bm{o}_t,s_t)}\nabla_{\psi}\bm{\pi}_{\theta,\psi}(\bm{o}_t,s_t),
%\end{align}
where $Q_{\rm{tot}}$ is the joint action-value function, which is discussed in detail in Subsection~\ref{Qtot}.

\subsection{Mutual Value Function Information}\label{Qtot}
As discussed in Subsection~\ref{IGM-section}, the joint action-value function $Q_{\rm{tot}}$ can be factorized by $Q_i(o_i, u_i)$ according to the IGM condition, while the decomposition methods have been criticized for the limited representative ability of the joint action-value function.
Therefore, we propose a new optimization objective, which maximizes the joint action-value function $Q_{\rm{tot}}(\bm{o}_t,\bm{u}_t)$ by maximizing the individual action-value function $Q_{i}(o_t^i, u_t^i)$ and the cooperative action-value function $Q_{ij}(o_t^i, o_t^j, u_t^i, u_t^j)$
\begin{align}
& Q_{\rm{tot}}(\bm{o}_t,\bm{u}_t) = \sum_{i=0}^{n}Q_i(o_t^i,u_t^i)+\sum_{i=0}^{n}\sum_{j\ne i}Q_{ij}(o_t^i,o_t^j,u_t^i,u_t^j), \\
& \pi_i(\bm{o}_t) = \mathop{ \arg\max}_{u_i}\left\{Q_i(o_t^i,u_t^i)+\sum_{j\ne i}\mathbb{E}_{u_j\sim \bar{\pi}_j}\left[Q_{ij}(o_t^i,o_t^j,u_t^i,u_t^j)\right]\right\},
\end{align}
where $\bar{\pi}_j$ denotes the fixed policy of agent j.
We name the new optimization objective as mutual value function information.
Based on the new objective, the IGM condition in Subsection~\ref{IGM-section} is modified as
\begin{align}\label{IGM_Modified}
\mathop{\arg\max}_{\textbf{u}}Q^{\bm{\pi}}_{\rm{tot}}(\bm{o},\textbf{u}) = 
\left(                 %左括号
\begin{array}{c}   %该矩阵一共3列，每一列都居中放置
\mathop{\arg\max}_{u_1}\left\{Q_1(o_1,u_1)+\sum_{j\ne 1}\mathbb{E}\left[Q_{1j}\right]\right\} \\  %第一行元素
\vdots\\  %第二行元素
\mathop{\arg\max}_{u_n}\left\{Q_n(o_n,u_n)+\sum_{j\ne n}\mathbb{E}\left[Q_{nj}\right]\right\}\\
\end{array}
\right).                 %右括号
\end{align}
However, accurately calculating the mutual value function is extremely hard due to the uncertainty of the cooperative relationship between agents, which is the core problem of MARL. 
We adopt the shared attention mechanism to approximately meet the condition in Equation~\ref{IGM_Modified}.
The individual action-value function $Q_i^{\phi}(\bm{o}_t,\bm{u}_t)$ is calculated according to all observation-action information of agents
\begin{eqnarray}
Q_i^{\phi}(\bm{o}_t,\bm{u}_t) = f_i(g_i(o^i_t,u^i_t),x^i_t), \\ 
x^i_t = \sum_{j\ne i}\alpha^j_t h(Vg_j(o^j_t,u^j_t)),
\end{eqnarray}
where $f_i,g_i,h$ denote multi-layer perceptions and $x_i$ denotes the weighted sum of per-agent action-values.
\begin{algorithm}[t]
	\label{algorithm_SAQ}
	\caption{Interactive Actor-Critic~(IAC)}
	\begin{algorithmic}[1]%一行一个标行号
		\STATE Initialize parameter vectors $\theta,\psi,\vartheta,\phi$
		\STATE Initialize replay buffer $\mathcal{D}$, parameters update rate $T$, training times $t$
		\FOR{each episode}
		\STATE Reset environment and receive $s$, $\bm{o}$
		\FOR{each step in episode}
		\STATE Select actions $\bm{u} \sim \bm{\pi}(\cdot \mid \bm{o},s)$ according to~(\ref{u_t})
		\STATE Send $\bm{u}$ to environment and receive $\bm{o}'$, $s'$, $r$
		\STATE Store transitions in replay buffer $\mathcal{D}$
		\IF{|$\mathcal{D}$| > $N_{\rm{Batch}}$}
		\STATE Sample minibatch $B$
		\STATE Calculate $\bm{u}\sim \bm{\pi}_{\theta}(\bm{o}_B,s_B)$ according to~(\ref{u_t})
		\STATE Calculate $Q_{\rm{tot}}(\bm{o}_B,\bm{u}, s_B;\vartheta, \phi)$ according to~(\ref{Q_tot})
		\STATE Update the joint-policy according to~(\ref{collaborative exploration})
		\STATE Calculate $Q_{\rm{tot}}(\bm{o}_B,\bm{u}_B,s_B;\vartheta,\phi)$ according to~(\ref{Q_tot})
		\STATE Calculate $\bm{u}'\sim \bm{\pi}_{\theta^-}(\bm{o}'_B,s_B)$ according to~(\ref{u_t})
		\STATE Calculate $Q_{\rm{tot}}(\bm{o}'_B,\bm{u}',s'_B;\vartheta^-,\phi^-)$ according to~(\ref{Q_tot})
		\STATE Update critics using $\nabla L_Q(\vartheta, \phi)$ according to~(\ref{Critic})
		\ENDIF
		\IF{t \% T = 0}
		\STATE Update target parameters:\\
		$\theta^- \leftarrow \theta$, $\psi^- \leftarrow \psi$, $\vartheta^- \leftarrow \vartheta$, $\phi^- \leftarrow \phi$ \\
		\ENDIF
		\ENDFOR
		\ENDFOR
	\end{algorithmic}
\end{algorithm}
The matrix $V$ is shared across agents. 
Individual action-value functions are parameterized by $\phi=\langle \phi_1, \phi_2, \dots,\phi_n \rangle$. 
We apply the query-key system to compute the attention weight $\alpha_j$ by comparing the embedding $e_j$
\begin{eqnarray}
\alpha_j \propto \exp\left(e_j^TW_k^TW_qe_i\right),
\end{eqnarray}
where $e_i=g_i(o_i,u_i)$, $W_k$ transforms $e_j$ into a key and $W_q$ transforms $e_i$ into a query. 
As for the implementation issue, we adopt a multi-head attention mechanism to select the different weighted mixture of per-agent action-values. 
The same mixing network as QMIX is adopted to calculate the joint action-value function $Q_{\rm{tot}}$
\begin{eqnarray}\label{Q_tot}
Q_{\rm{tot}}(\bm{o}_t,\bm{u}_t,s_t;\vartheta,\phi)= \text{MIX} \left(Q_1^{\phi}(\bm{o}_t,\bm{u}_t),...,Q_n^{\phi}(\bm{o}_t,\bm{u}_t),s_t;\vartheta\right),
\end{eqnarray}
where MIX denotes the mixing network parameterized by $\vartheta$. 
The attention-based joint action-value network is trained end-to-end by minimizing the following loss
\begin{align}\label{Critic}
L_Q(\vartheta,\phi) = \sum_{i=1}^{|b|}\left[\left(y_i^{\rm{tot}}-Q_{\rm{tot}}(\bm{o}_t,\bm{u}_t,s_t;\vartheta,\phi)\right)^2\right],\\
y^{\rm{tot}}=r+\gamma Q_{\rm{tot}}(\bm{o}_{t+1},\bm{u}_{t+1},s_{t+1};\vartheta^-,\phi^-)\mid_{\bm{u}_{t+1}=\bm{\pi}^-(\bm{o}_{t+1},s_{t+1})},
\end{align}
where $b$ is the batch sampled from the replay buffer and $\bm{\pi}^-$ denotes the target joint stochastic policy. 
The target mixing network, individual action-value network are parameterized respectively by $\vartheta^-$, $\phi^-$.

%%%%%%%%%%%%%%%%%%%%%%%%%%%%%%%%%%%%%%%%%%%%%%%%%%%%%%%%%%%%%%%%%%%%%%%%

\section{Experiments}\label{Experiment result}
To evaluate our algorithm, we design a series of multi-agent cooperative experiments to compare IAC with the state-of-the-art MARL algorithms.
Hyper-parameters and implementation details are listed in Subsection~\ref{setup} and Appendix~\ref{Experimental Details}.
The state-of-the-art MARL algorithms are represented in Subsection~\ref{Baseline}.
The multi-agent cooperative experiments are shown in Appendix~\ref{Multi-agent Environments}.
The performance comparison between different methods and detailed analysis are discussed in Subsection~\ref{results}.
\subsection{Setting}\label{setup}
\begin{figure}[t]
    \centering
	\subfloat[DDPG-MIX.]
	{
		\includegraphics[width=1.0in]{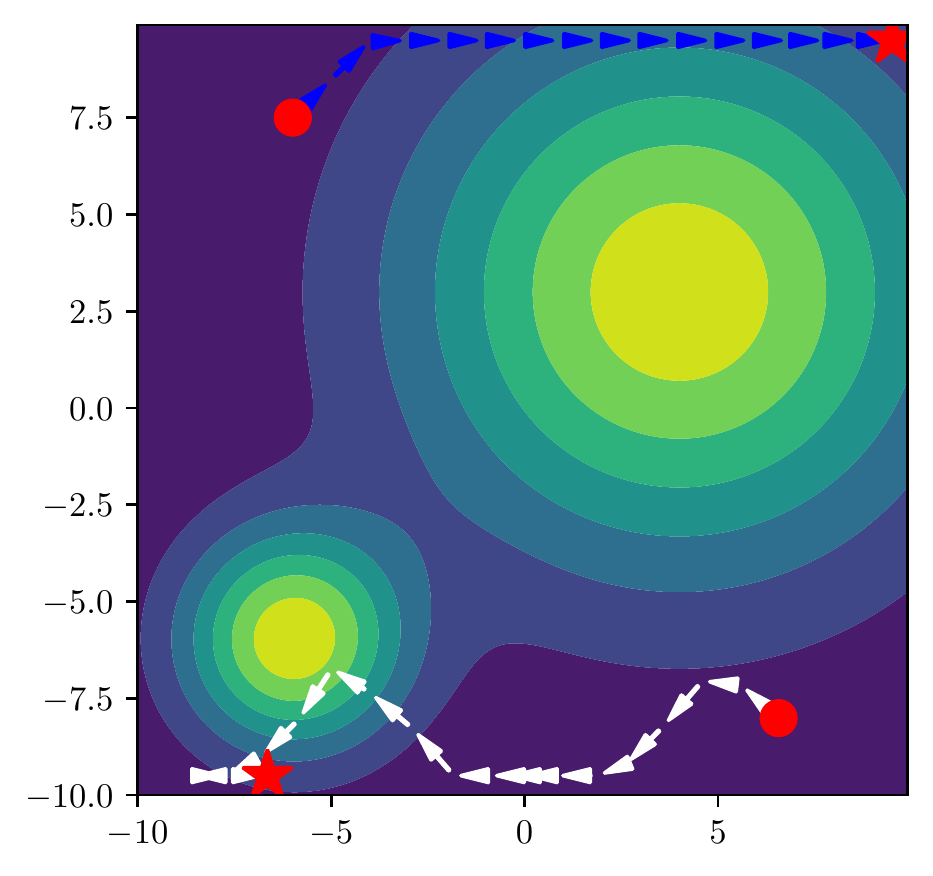}
	}
	\subfloat[Attn-QMIX.]
	{
		\includegraphics[width=1.0in]{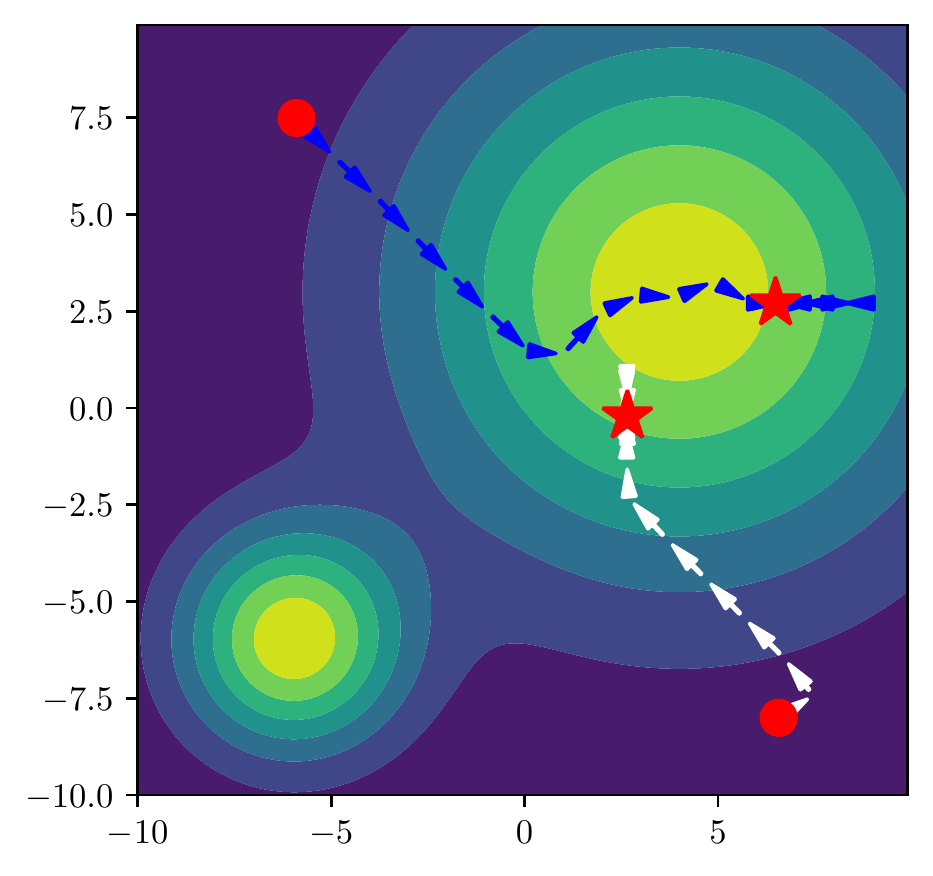}
	}
	\subfloat[IAC.]
	{
		\includegraphics[width=1.0in]{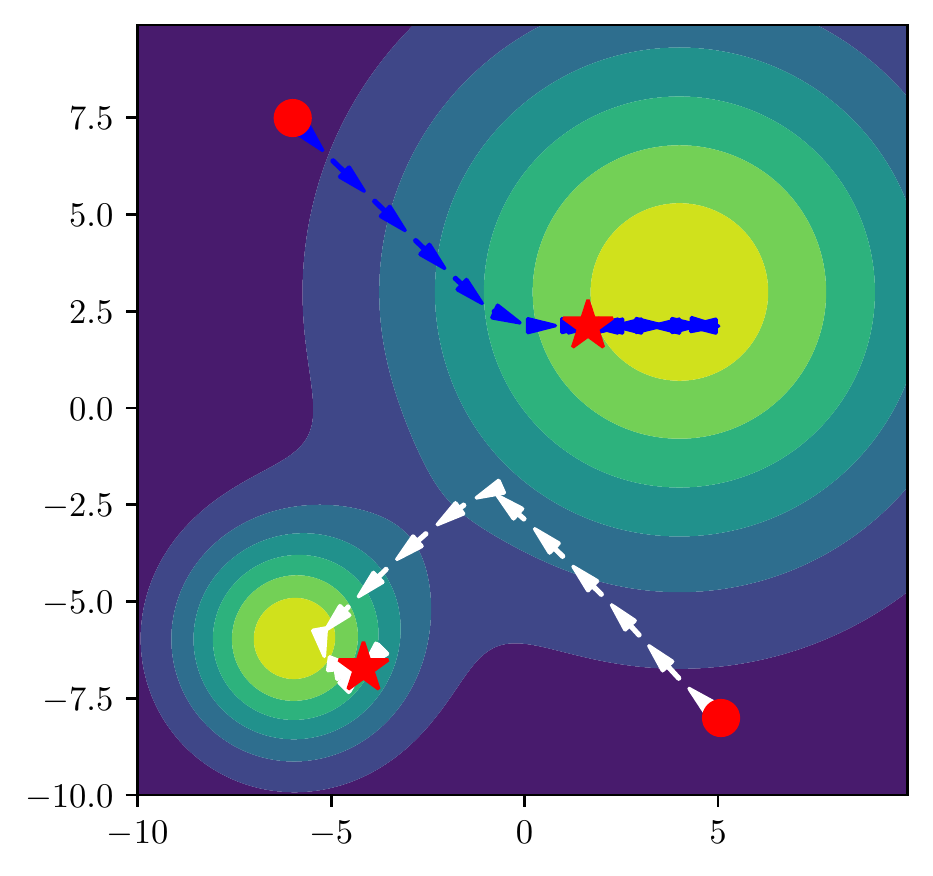}
	}
    \caption{Visualization of  value distributions and trajectories of various methods.
    Dotted lines denote trajectories generated by different algorithms.
    The red circle is the starting point and the red star is the ending point.}
    \label{trajectory}
\end{figure}

\begin{figure}[t]
	\centering
	\subfloat[Independent policy distribution.]
	{
		\includegraphics[width=1.4in]{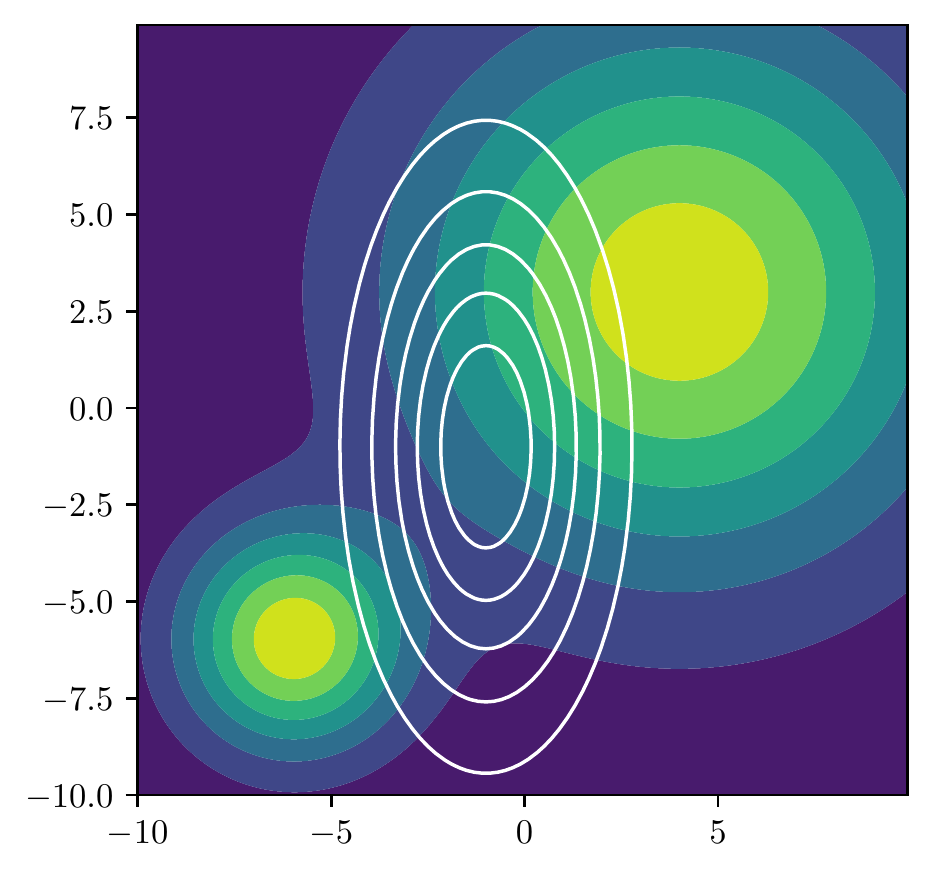}
	}\hspace{+1mm}
	\subfloat[Collaborative policy distribution.]
	{
		\includegraphics[width=1.4in]{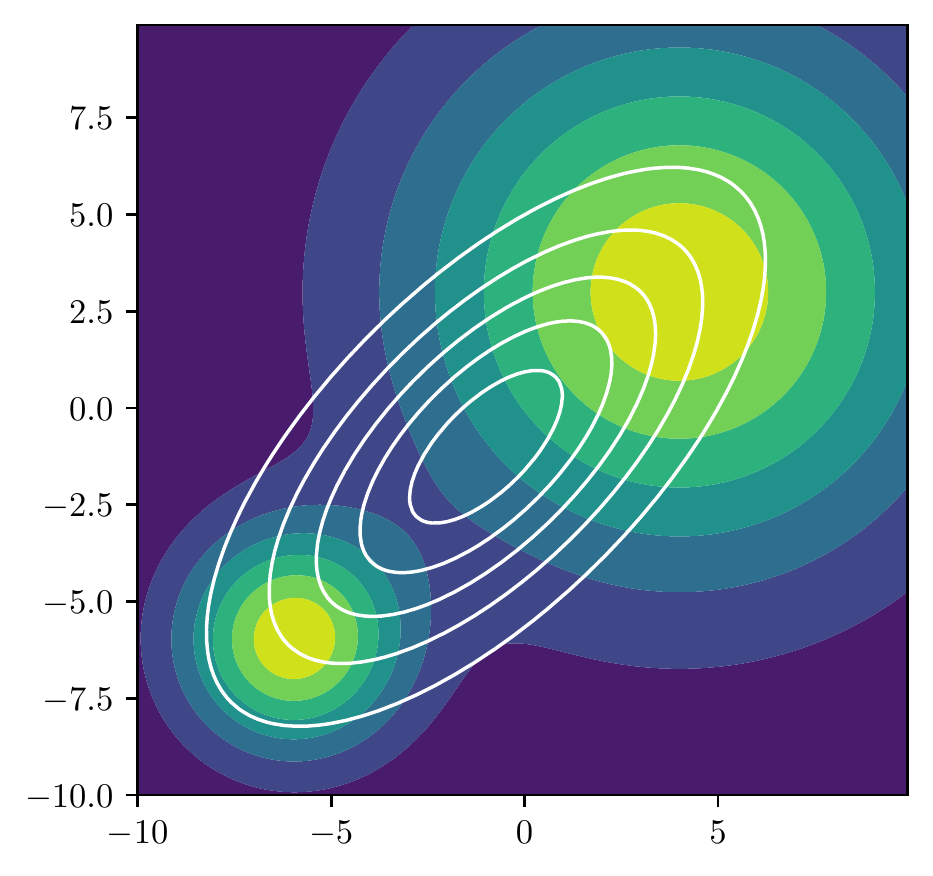}
	}
	\caption{Visualization of value distributions and policy distributions. 
	Colored area denotes independent value distributions. 
	Bright color area indicates high reward.
	White circle on (a) represents the independent policy distribution. 
	White circle on (b) represents the collaborative policy distribution.}
	\label{MGaussianD}
\end{figure}

$\textbf{Simple World}$. We design a task named \textit{Simple World} to evaluate the exploration ability of the collaborative exploration module and representative ability of Attention-based QMIX.
Simple World is a two-player one-action one-step task, which has no state and will return the reward according to taken actions.
For convenience, we adopt two independent Gaussian distribution to represent the value distributions.
The means are $[4,3]$ and $[-6,-6]$. The variances are 5 and 2 respectively.
There two value distributions denote the reward distributions in the task.

\noindent $\textbf{Multi-Pendulum}$. In this subsection, we extend the pendulum task from single-agent to multi-agent tasks. Single-pendulum in the gym is a classic benchmark. 
In detail, we aim at swinging up a pendulum starting from a random position and keeping it upright. 
Compared with the single-pendulum task, the multi-agent classic control task has multiple pendulums to be controlled. 
Each pendulum only receives the self-observation and the shared reward, which indicates the multi-pendulum environment is a Dec-POMDP task.
Moreover, the reward information is sparse in the environment.
When all angles of pendulums are less than 0.2, the immediate reward is +5. 
Otherwise, all agents receive reward 0. 

\noindent $\textbf{Predator-prey}$ is a classic multi-agent cooperative benchmark.
In the predator-prey game, $N$ slower cooperating agents chase a faster adversary around a randomly generated environment with $L$ large landmarks. 
We set three cooperating agents, two landmarks and one adversary. 
The adversary moves randomly. 
When the cooperative agents collide with the adversary, these cooperative agents receive a shared reward +10. 
Otherwise, the immediate reward is 0. 
Predator-prey is a Dec-MDP task as agents can observe the relative positions and velocities of other agents and landmarks at all times.

\noindent $\textbf{PO-Predator-prey}$ is an enhanced version of the predator-prey. 
Different from the predator-prey task, agents in the PO-Predator-prey task can only receive the information within the distance of 0.8. 
Other settings, such as the number of agents and the reward function, are the same as the predator-prey task.

\begin{figure*}[t]
	\centering
	\subfloat[Average rewards in multi-pendulums environment.]
	{
	    \includegraphics[width=2.1in]{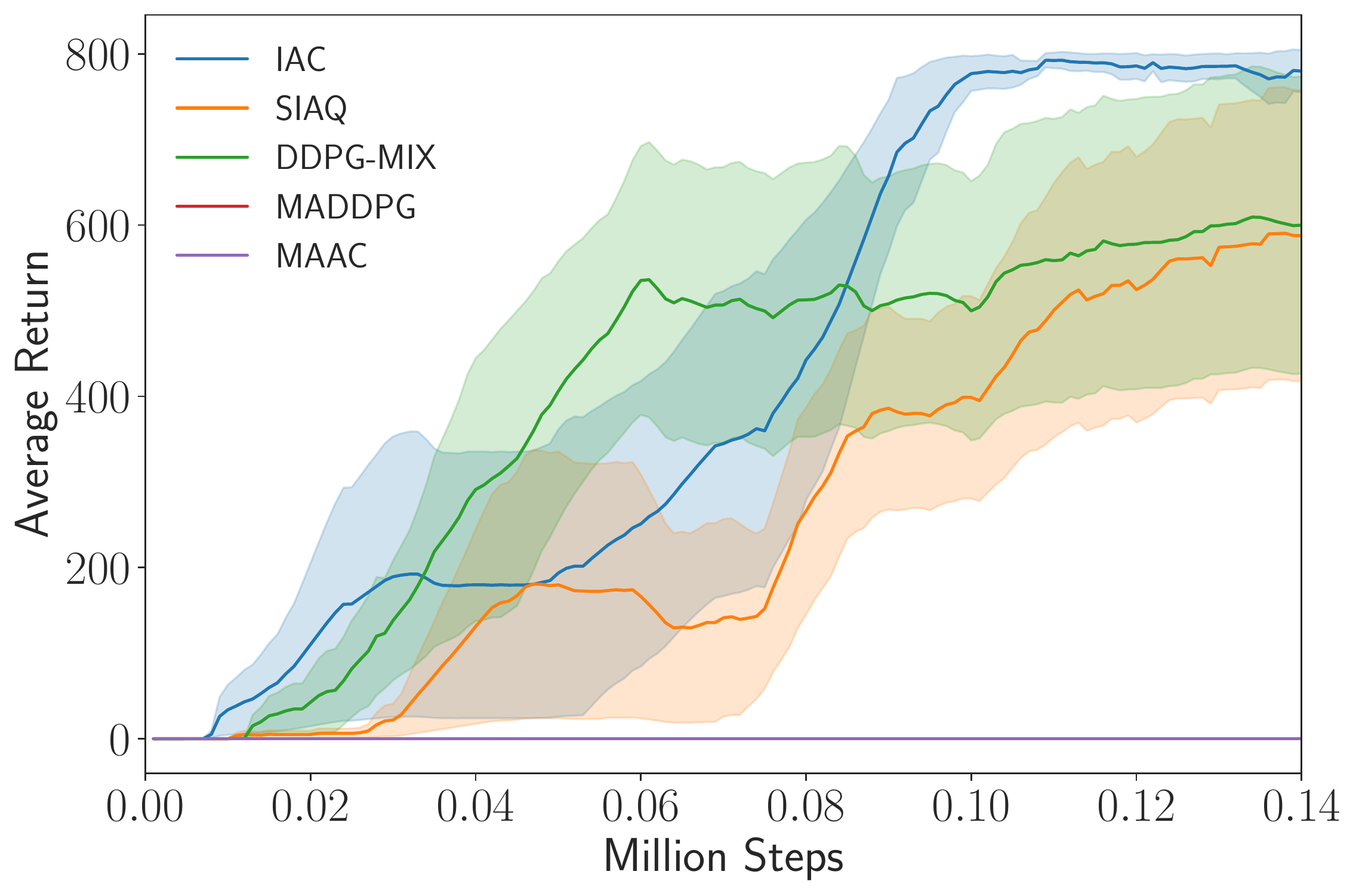}
	}
	\subfloat[Average rewards in Predator-prey task.]
	{
		\includegraphics[width=2.1in]{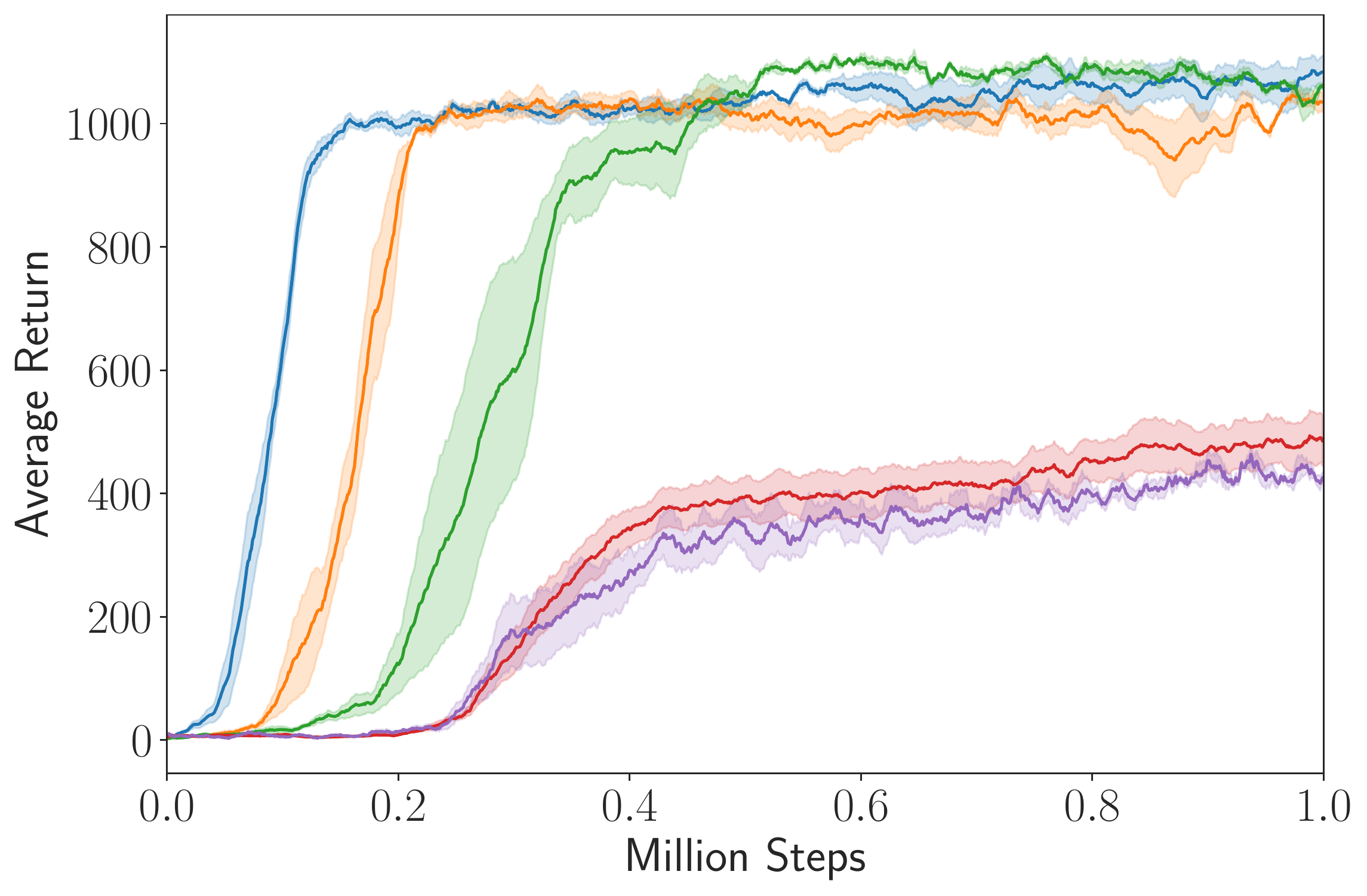}
	} \label{1}
	\subfloat[Average rewards in PO-Predator-prey task.]
	{
		\includegraphics[width=2.1in]{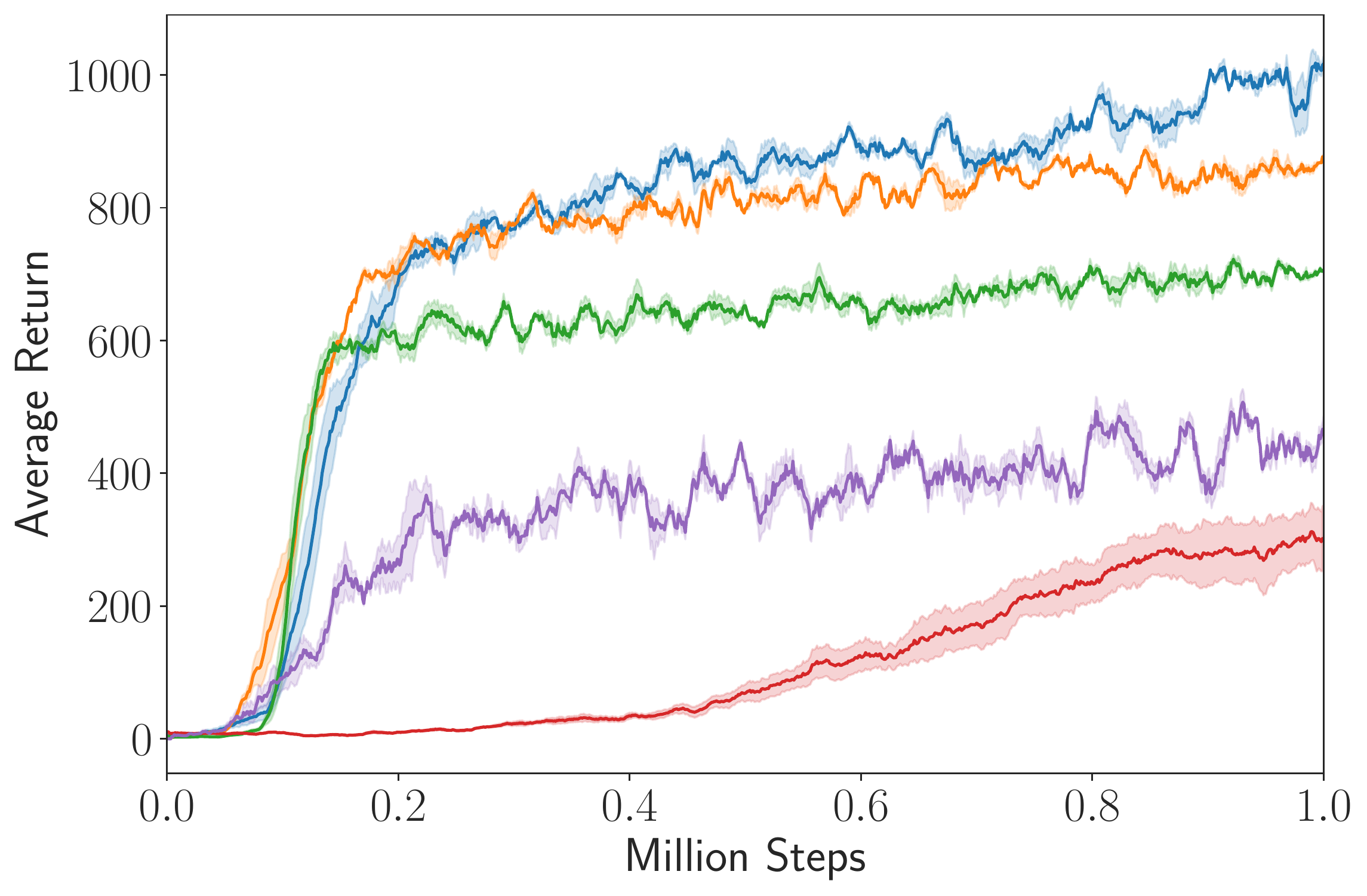}
	}
	\caption{Evaluation of various methods in Classic Control, Predator-prey and POMDP-predator-prey tasks. 
	IAC achieves better performance in all tasks compared with other baselines.}
	\label{predator-prey-f}
\end{figure*}

\subsection{Baselines}\label{Baseline}
We compare our algorithms with the state-of-the-art approaches: MADDPG and MAAC, which are decentralized policies of centralized training.
To ensure the accuracy of the reproduction, we apply the official code\footnote{ https://github.com/openai/maddpg, https://github.com/shariqiqbal2810/MAAC}.

Since improvements of IAC come from both actor and critic modules, we set different baselines to evaluate their performances respectively. 
We combine the value decomposition methods and the multi-agent deterministic policy.
This basic approach is named \textbf{DDPG-MIX}. We replace respectively the exploration policy and attention module in IAC to generate the corresponding baselines: Soft-QMIX~(\textbf{SQ}) and Attention-QMIX~(\textbf{Attn-QMIX}). 

As a comparison for maximum entropy policy, a naive method to solve the problem in Subsection~\ref{CE Module} is defining the soft individual action-value function:
\begin{eqnarray}
Q_i = \sum_{t=0}^{T}\mathbb{E}_{(o_t^i, u_t^i)\sim \tau_{\pi_i}}\left[r(o_t^i, u_t^i) + \alpha \mathcal{H}(\pi_i(\cdot \mid o_t^i))\right].
\end{eqnarray}
Mixing Network in Section~\ref{Qtot} calculates the joint action-value function $Q_{\rm{tot}}$ by approximating weight $\vartheta$ in neurual network
\begin{align}
Q_{\rm{tot}} &= \sum_{i=1}^{n}\text{MIX}\left(Q_i(o_t^i, u_t^i); \vartheta \right) \notag \\
&= \sum_{i=1}^{n}\text{MIX}\left(\sum_{t=0}^{T}\mathbb{E}_{(o_t^i, u_t^i)\sim \tau_{\pi_i}}\left[r(o_t^i, u_t^i) + \alpha \mathcal{H}(\pi_i(\cdot \mid o_t^i))\right]; \vartheta\right),
\end{align}
where $\rm{MIX}$ denotes the Mixing Network, such as the QMIX Network. 
Based on individual policies, the entropy of the joint stochastic policy is $\sum_{i=1}^{n}\text{MIX}\left(\sum_{t=0}^{T}\mathbb{E}_{(o_t^i, u_t^i)\sim \tau_{\pi_i}}\left[\mathcal{H}(\pi_i(\cdot \mid o_t^i))\right];\vartheta \right)$. 
We name this method as Soft Individual QMIX~(\textbf{SIQ}), which has the similar critic structure as IAC but different exploration module. Moreover, we adopt the same attention-based critic module in SIQ to generate the corresponding baseline: Soft-Individual-Attention-QMIX~(\textbf{SIAQ}). 
All the above methods use the same hyperparameters and networks as IAC.

% However, QMIX network has a limited ability due to its monotonic structural constraints, which ignores some important information~\cite{wang2020qplex}. 

\subsection{Results}\label{results}
\begin{figure}[t]
	\centering
	\subfloat[Visualization at the 1st step]{
		\includegraphics[width=3.5cm]{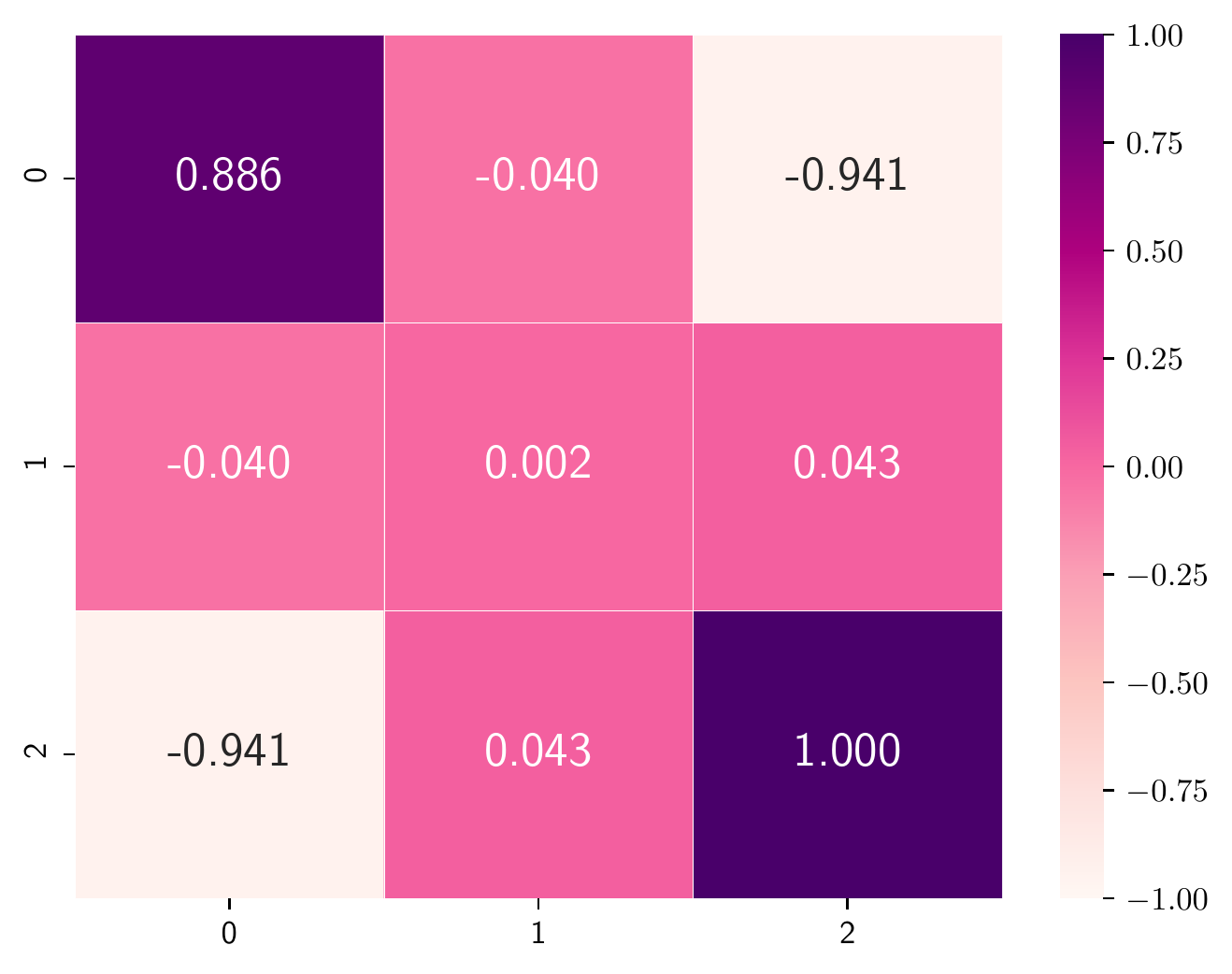}
	}
	\subfloat[Visualization at the 20th step]{
		\includegraphics[width=3.5cm]{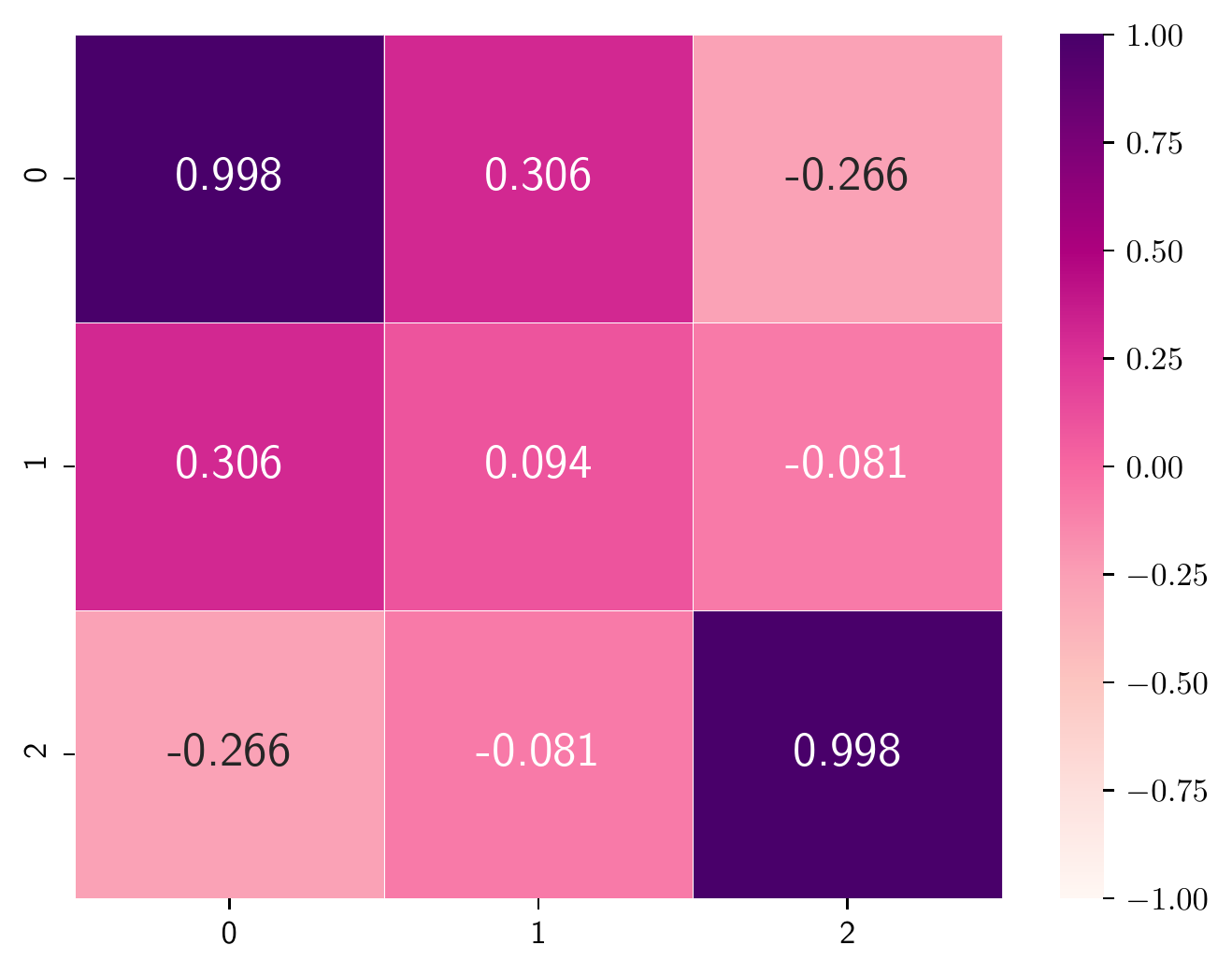}
	}

	\subfloat[Visualization at the 40th step]{
		\includegraphics[width=3.5cm]{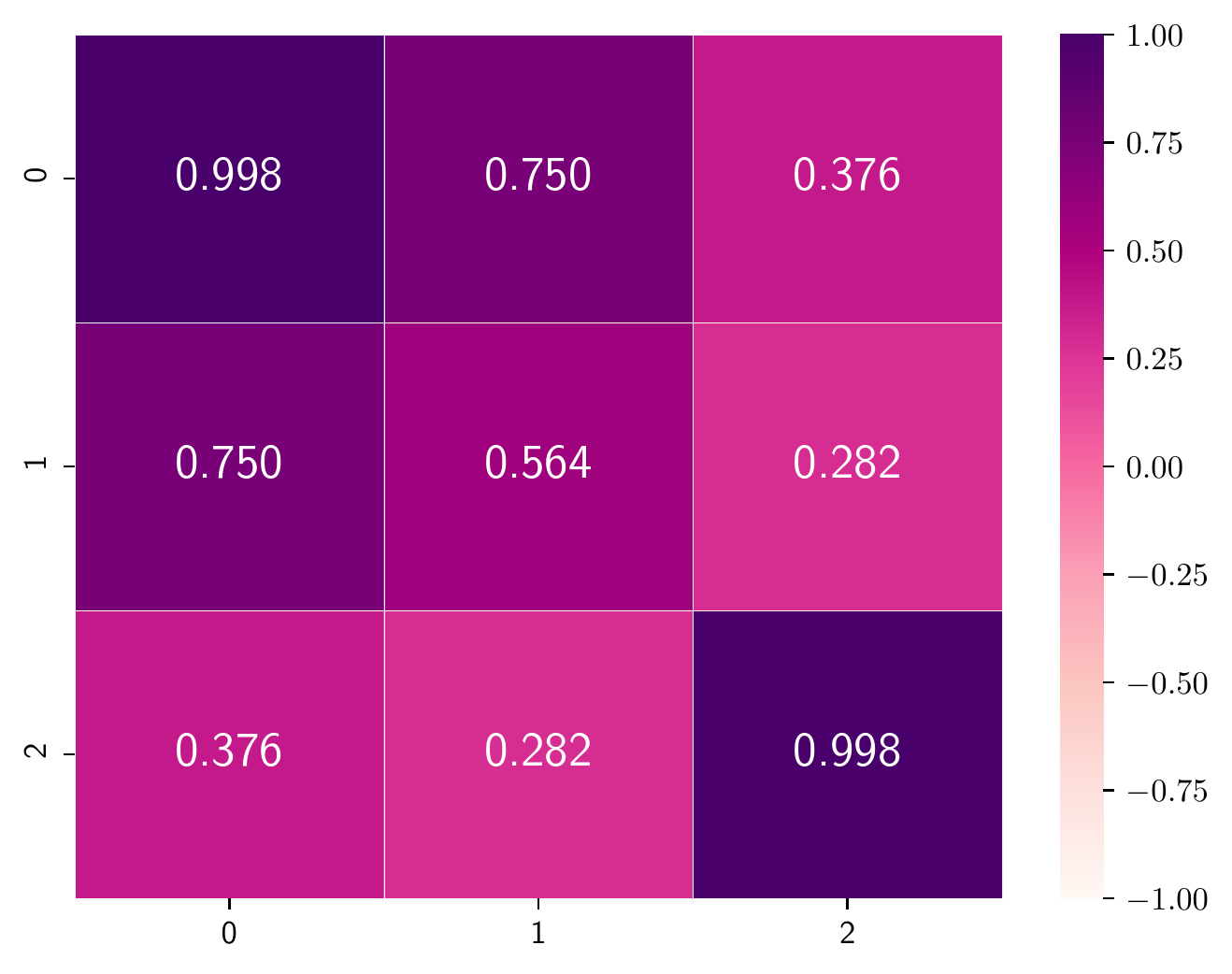}
	}
	\subfloat[Visualization at the 60th step]{
		\includegraphics[width=3.5cm]{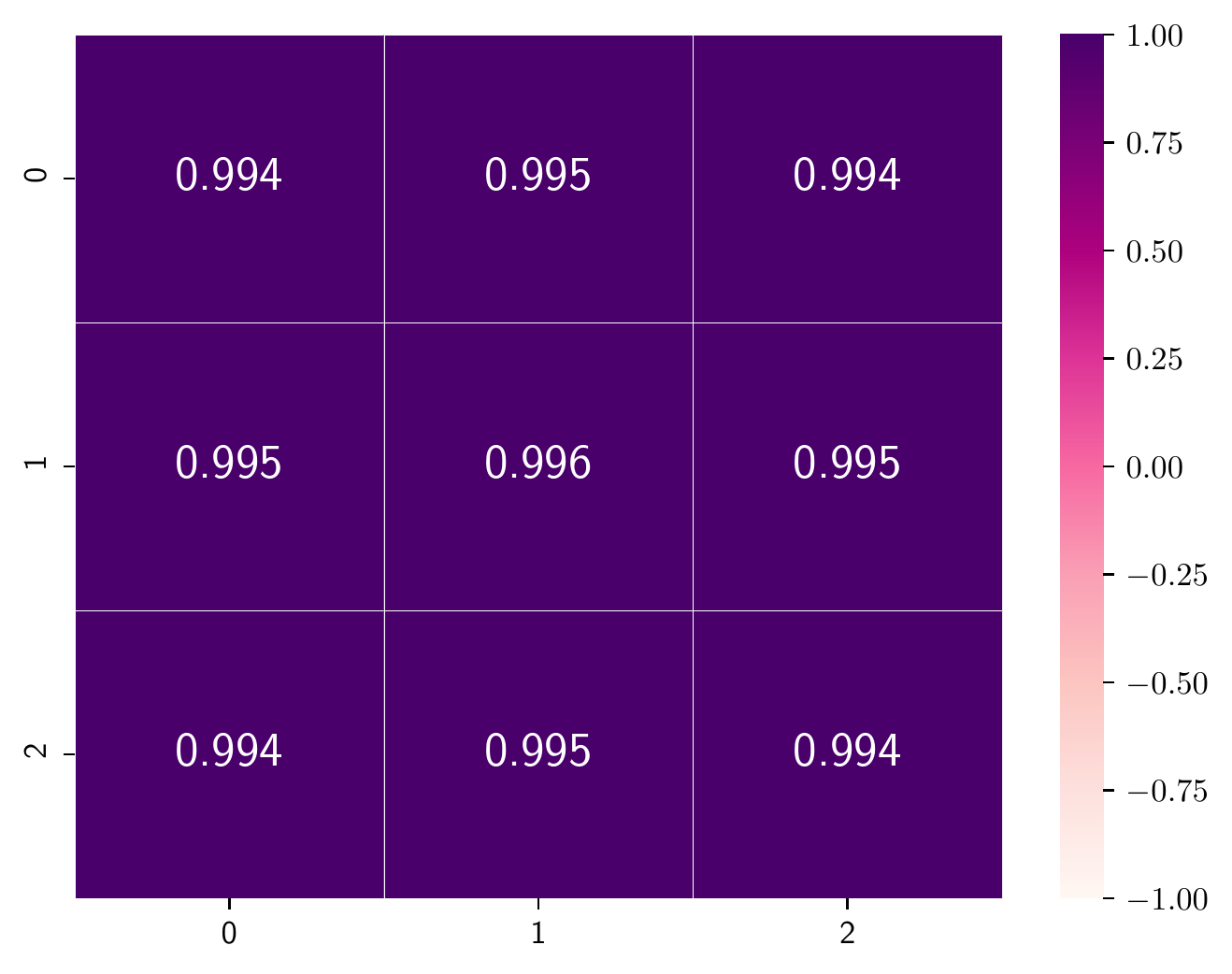}
	}
	\caption{Visualization of the covariance matrix in IAC at different steps in the multi-pendulum task. 
	We set the number of the pendulums is 3.
	The covariance matrix indicates the relationship of behavior strategies. Values in the covariance matrix change from irregular to regular, which indicates that there is a clear relationship between agents.}
	\label{k-outs}
\end{figure}

%\begin{figure}[t]
%	\centering
%	\subfloat[screenshot at 70th step]{
%		\includegraphics[width=4.5cm]{3pendulum_1.png}
%	}
%	\subfloat[screenshot at 80th step]{
%		\includegraphics[width=4.0cm]{3pendulum_2.png}
%	}
%	\caption{The performance of the trained model in multi-pendulum environment. We take screenshots at random time steps and we found that the strategies are similar.}
%	\label{pendulum-2}
%\end{figure}

$\textbf{Simple World}$.
We evaluate DDPG-MIX, Attn-QMIX and IAC in the task, which is illustrated in the Figure~\ref{trajectory}. 
Experimental results indicate that the Attn-QMIX has a stronger representative ability compared with QMIX. 
Moreover, compared with the individual exploration, the collaborative exploration module in IAC promotes to explore the more meaningful area.
The experimental result can be explained visually in Figure~\ref{MGaussianD}, where we consider two different exploration policies: the independent policy and the collaborative policy.
Both the independent policy and collaborative policy are denoted by a multivariate Gaussian distribution with the same mean $[-1, -1]$. 
The covariance matrix of the independent policy is $\Sigma=\left[4,0;0,20\right]$. 
The covariance matrix of the collaborative policy is $\Sigma=\left[15,10;10,15\right]$.
The independent policy distribution has some meaningless areas, which induces useless exploration. 
On the contrary, the collaborative policy distribution prefers to explore high reward areas, which indicates that the collaborative exploration is more effective compared with the independent exploration. 

\noindent $\textbf{Multi-Pendulum}$. The experimental results are shown in Figure~\ref{predator-prey-f}. 
The improvements of IAC, SIAQ, and DDPG-MIX are apparent compared with MADDPG and MAAC, which indicates the superiority of value decomposition methods. We visualize the covariance matrix $\Sigma$ in IAC at different training steps, which is shown in Figure~\ref{k-outs}.
At the 1st step, the covariance matrix consists of irregular values.
The policy of the first agent is positively related to itself and negatively related to the third agent. 
Gradually, the relationship of agent policies is positive.
Finally, the relationship of agent policies is stable. 
In the multi-pendulum task, agents have the same objective and their policies should be positively related, which has been satisfied by the visualization result.

\begin{figure}[t]
	\centering
	\includegraphics[width=2.9in]{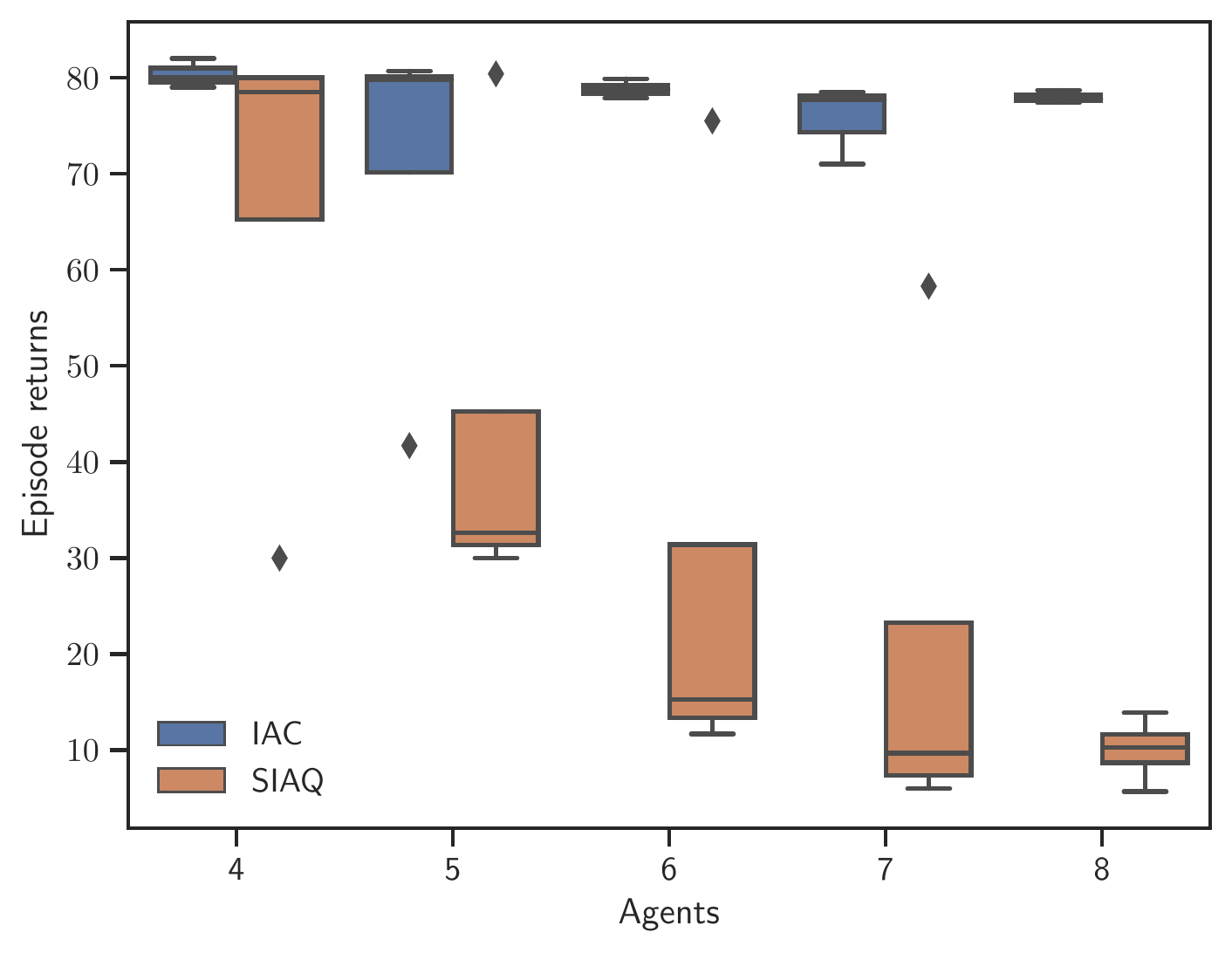}
	\caption{Scalability performance in the multi-pendulum task. 
	When all angles of the pendulums are less than 0.2, the immediate reward is +0.5. 
	When there are one or three pendulums meeting the angle requirement, the immediate rewards are +0.1 and +0.2 respectively.}
	\label{box}
\end{figure}

Moreover, we evaluate the scalability of IAC and SIAQ in multi-pendulum task.
As the number of agents grows, rewards in this task are extremely sparse.
To reduce the exploration difficulty, agents receive few rewards when a few pendulums meet the angle requirement, which is illutrated in Figure~\ref{box}. 
Experimental results illustrate that the performance of IAC does not deteriorate when pendulums are increased, while the performance of SIAQ is not scalable.

\noindent $\textbf{Predator-prey and PO-Predator-prey}$. Various algorithms are evaluated in Predator-prey and PO-Predator-prey tasks, whose results are illustrated in the Figure~\ref{predator-prey-f}. 
Experimental results indicate that value-based methods achieve better performance than MADDPG and MAAC. 
Moreover, collaborative exploration module has superior performance than the deterministic policy. 
Compared with other methods, our proposed algorithm~(IAC) is competitive in both environments. As the reward setting in the Predator-prey task is similar to the multi-pendulum task, MADDPG and MAAC have limited capacity to find the optimal individual action-value function for each agent. 
We set the average reward threshold to 1000, where three predators can succeed to capture prey cooperatively. 
Compared with DDPG-MIX and SIAQ, IAC saves 0.295 and 0.06 million steps respectively to complete the predator-task.

\begin{figure}[t]
	\centering
	
	\subfloat[Saved Steps.]
	{
	 \includegraphics[width=1.5in]{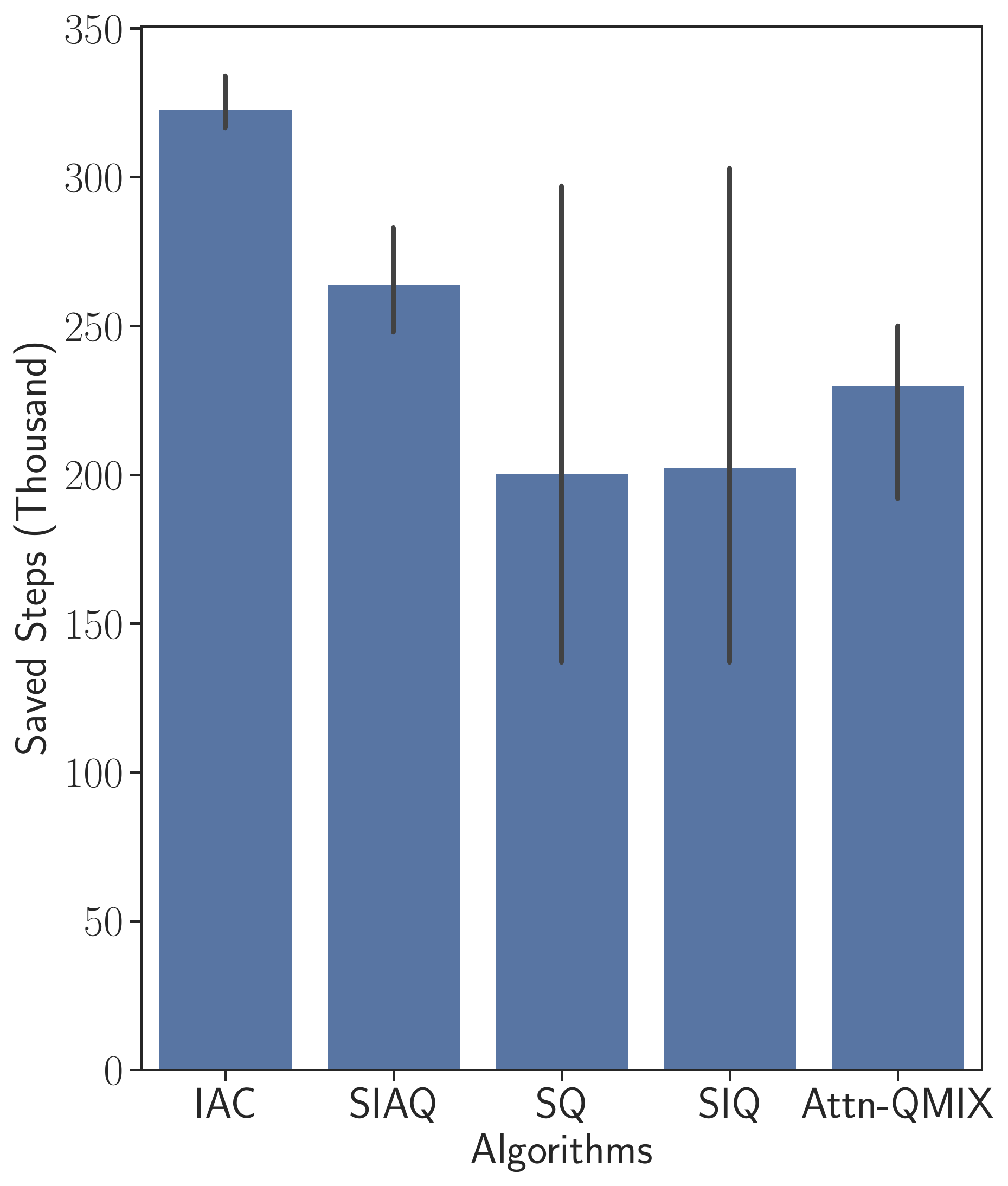}
	}
	\subfloat[Diff-Return.]
	{
	 \includegraphics[width=1.5in]{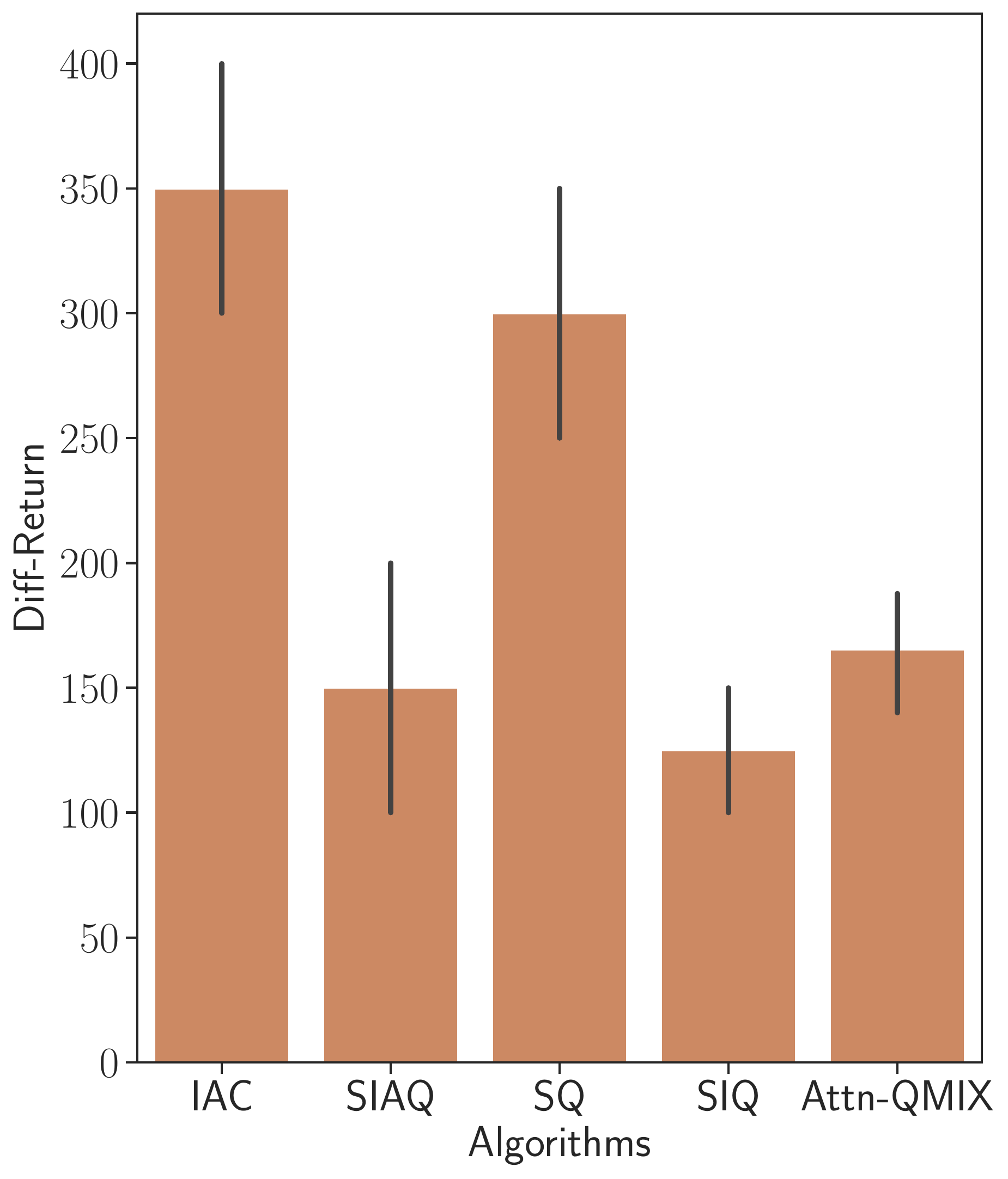}
	}
	\caption{The performance comparison between IAC, SIAQ and baselines. 
	Assume agents controlled by DDPG-MIX capture prey cooperatively in the Predator-prey task at time $t_1$.
    Other methods achieve the same objective at time $t_i$.
    The saved steps is $\Delta t = t_1 - t_i$, and the unit is one thousand step. 
	The diff-return is the episode return difference between IAC, SIAQ and DDPG-MIX in the PO-Predator-prey task.}
	\label{zf}
\end{figure}

In the PO-Predator-prey task, agents only receive the local information within 0.8 distance.
Experimental results indicate that the performance of MADDPG, DDPG-MIX and SIAQ have decreased. 
As the MAAC applies the attention mechanism to calculate the individual action-value function, the performance of MAAC is maintained. 
Compared with these methods, IAC still achieves better robustness and succeeds to capture prey cooperatively. 
Since IAC and SIAQ have been modified on the policy and value-function estimation sides, the respect performance analysis is necessary, which is represented in Figure~\ref{zf}. 
Compared with SQ and SIQ, IAC and SIAQ achieve better performance in terms of the saved time and episode reward.
Therefore, both the collaborative exploration policy and shared attention mechanism have a positive influence on the performance. 

%We also visualize the covariance matrix $\Sigma$ for the SAQ algorithm and the results are illustrated in Appendix.~\ref{visualization}.

%%%%%%%%%%%%%%%%%%%%%%%%%%%%%%%%%%%%%%%%%%%%%%%%%%%%%%%%%%%%%%%%%%%%%%%%

\section{Conclusion}
In this paper, we propose a novel cooperative MARL algorithm named IAC to model the interaction between agents. 
IAC adopts the collaborative exploration module for the policy to couple the individual policies and the joint-policy of agents, which is trained by maximizing the entropy-regularized expected return.
For the value estimation, IAC utilizes a shared attention mechanism to estimate the mutual value function, which considers the impact of the teammates.
Moreover, IAC extends the value decomposition methods to continuous control tasks. 
Taking the advantages for both sides of collaborative exploration and mutual value function, IAC not only outperforms existing baselines but also achieves superior cooperation, robustness and scalability.

In the future, we will extend our work to large-scale multi-agent tasks liking MAgents~\cite{yang2018mean,ganapathi2020multi}, because AI multi-agent system gradually contains hundreds of agents, such as the Warehouse Robot System and Bee colony system.
These large-scale multi-agent systems have more complex interaction and tasks.
We believe that solving modeling interaction in multi-agent systems would be one-step toward Multi-Agent AI.

%%% The acknowledgments section is defined using the "acks" environment
%%% (rather than an unnumbered section). The use of this environment 
%%% ensures the proper identification of the section in the article 
%%% metadata as well as the consistent spelling of the heading.

\begin{acks}
This work was funded by the National Key Research and Development Project of China under Grant 2017YFC0704100 and Grant 2016YFB0901900, in part by the National Natural Science Foundation of China under Grant 61425027, the 111 International Collaboration Program of China under Grant BP2018006, and BNRist Program (BNR2019TD01009), and the National Innovation Center of High Speed Train R\&D project (CX/KJ-2020-0006).
\end{acks}
 
%%%%%%%%%%%%%%%%%%%%%%%%%%%%%%%%%%%%%%%%%%%%%%%%%%%%%%%%%%%%%%%%%%%%%%%%

%%% The next two lines define, first, the bibliography style to be 
%%% applied, and, second, the bibliography file to be used.
\clearpage

\bibliographystyle{ACM-Reference-Format} 
\balance
\bibliography{sample}

%%% -*-BibTeX-*-
%%% Do NOT edit. File created by BibTeX with style
%%% ACM-Reference-Format-Journals [18-Jan-2012].

\begin{thebibliography}{43}

%%% ====================================================================
%%% NOTE TO THE USER: you can override these defaults by providing
%%% customized versions of any of these macros before the \bibliography
%%% command.  Each of them MUST provide its own final punctuation,
%%% except for \shownote{}, \showDOI{}, and \showURL{}.  The latter two
%%% do not use final punctuation, in order to avoid confusing it with
%%% the Web address.
%%%
%%% To suppress output of a particular field, define its macro to expand
%%% to an empty string, or better, \unskip, like this:
%%%
%%% \newcommand{\showDOI}[1]{\unskip}   % LaTeX syntax
%%%
%%% \def \showDOI #1{\unskip}           % plain TeX syntax
%%%
%%% ====================================================================

\ifx \showCODEN    \undefined \def \showCODEN     #1{\unskip}     \fi
\ifx \showDOI      \undefined \def \showDOI       #1{#1}\fi
\ifx \showISBNx    \undefined \def \showISBNx     #1{\unskip}     \fi
\ifx \showISBNxiii \undefined \def \showISBNxiii  #1{\unskip}     \fi
\ifx \showISSN     \undefined \def \showISSN      #1{\unskip}     \fi
\ifx \showLCCN     \undefined \def \showLCCN      #1{\unskip}     \fi
\ifx \shownote     \undefined \def \shownote      #1{#1}          \fi
\ifx \showarticletitle \undefined \def \showarticletitle #1{#1}   \fi
\ifx \showURL      \undefined \def \showURL       {\relax}        \fi
% The following commands are used for tagged output and should be
% invisible to TeX
\providecommand\bibfield[2]{#2}
\providecommand\bibinfo[2]{#2}
\providecommand\natexlab[1]{#1}
\providecommand\showeprint[2][]{arXiv:#2}

\bibitem[\protect\citeauthoryear{Ahmed, Le~Roux, Norouzi, and Schuurmans}{Ahmed
  et~al\mbox{.}}{2019}]%
        {ahmed2019understanding}
\bibfield{author}{\bibinfo{person}{Zafarali Ahmed}, \bibinfo{person}{Nicolas
  Le~Roux}, \bibinfo{person}{Mohammad Norouzi}, {and} \bibinfo{person}{Dale
  Schuurmans}.} \bibinfo{year}{2019}\natexlab{}.
\newblock \showarticletitle{Understanding the impact of entropy on policy
  optimization}. In \bibinfo{booktitle}{\emph{International Conference on
  Machine Learning}}. PMLR, \bibinfo{pages}{151--160}.
\newblock


\bibitem[\protect\citeauthoryear{Al~Mallah, Quintero, and Farooq}{Al~Mallah
  et~al\mbox{.}}{2019}]%
        {al2019cooperative}
\bibfield{author}{\bibinfo{person}{Ranwa Al~Mallah}, \bibinfo{person}{Alejandro
  Quintero}, {and} \bibinfo{person}{Bilal Farooq}.}
  \bibinfo{year}{2019}\natexlab{}.
\newblock \showarticletitle{Cooperative evaluation of the cause of urban
  traffic congestion via connected vehicles}.
\newblock \bibinfo{journal}{\emph{IEEE Transactions on Intelligent
  Transportation Systems}} \bibinfo{volume}{21}, \bibinfo{number}{1}
  (\bibinfo{year}{2019}), \bibinfo{pages}{59--67}.
\newblock


\bibitem[\protect\citeauthoryear{Bellemare, Srinivasan, Ostrovski, Schaul,
  Saxton, and Munos}{Bellemare et~al\mbox{.}}{2016}]%
        {bellemare2016unifying}
\bibfield{author}{\bibinfo{person}{Marc Bellemare}, \bibinfo{person}{Sriram
  Srinivasan}, \bibinfo{person}{Georg Ostrovski}, \bibinfo{person}{Tom Schaul},
  \bibinfo{person}{David Saxton}, {and} \bibinfo{person}{Remi Munos}.}
  \bibinfo{year}{2016}\natexlab{}.
\newblock \showarticletitle{Unifying count-based exploration and intrinsic
  motivation}. In \bibinfo{booktitle}{\emph{Advances in neural information
  processing systems}}. \bibinfo{pages}{1471--1479}.
\newblock


\bibitem[\protect\citeauthoryear{Bertsekas}{Bertsekas}{2020}]%
        {bertsekas2020multiagent}
\bibfield{author}{\bibinfo{person}{Dimitri Bertsekas}.}
  \bibinfo{year}{2020}\natexlab{}.
\newblock \showarticletitle{Multiagent Value Iteration Algorithms in Dynamic
  Programming and Reinforcement Learning}.
\newblock \bibinfo{journal}{\emph{arXiv preprint arXiv:2005.01627}}
  (\bibinfo{year}{2020}).
\newblock


\bibitem[\protect\citeauthoryear{Cassano, Yuan, and Sayed}{Cassano
  et~al\mbox{.}}{2018}]%
        {cassano2020multi}
\bibfield{author}{\bibinfo{person}{Lucas Cassano}, \bibinfo{person}{Kun Yuan},
  {and} \bibinfo{person}{Ali~H Sayed}.} \bibinfo{year}{2018}\natexlab{}.
\newblock \showarticletitle{Multi-Agent Fully Decentralized Value Function
  Learning with Linear Convergence Rates}.
\newblock \bibinfo{journal}{\emph{arXiv preprint arXiv:1810.07792}}
  (\bibinfo{year}{2018}).
\newblock


\bibitem[\protect\citeauthoryear{Cheng}{Cheng}{2019}]%
        {bahdanau2015neural}
\bibfield{author}{\bibinfo{person}{Yong Cheng}.}
  \bibinfo{year}{2019}\natexlab{}.
\newblock \showarticletitle{Semi-supervised learning for neural machine
  translation}.
\newblock In \bibinfo{booktitle}{\emph{Joint Training for Neural Machine
  Translation}}. \bibinfo{publisher}{Springer}, \bibinfo{pages}{25--40}.
\newblock


\bibitem[\protect\citeauthoryear{Chu, Chinchali, and Katti}{Chu
  et~al\mbox{.}}{2020}]%
        {chu2019multi}
\bibfield{author}{\bibinfo{person}{Tianshu Chu}, \bibinfo{person}{Sandeep
  Chinchali}, {and} \bibinfo{person}{Sachin Katti}.}
  \bibinfo{year}{2020}\natexlab{}.
\newblock \showarticletitle{Multi-agent Reinforcement Learning for Networked
  System Control}.
\newblock \bibinfo{journal}{\emph{arXiv preprint arXiv:2004.01339}}
  (\bibinfo{year}{2020}).
\newblock


\bibitem[\protect\citeauthoryear{Das, Gervet, Romoff, Batra, Parikh, Rabbat,
  and Pineau}{Das et~al\mbox{.}}{2019}]%
        {das2019tarmac}
\bibfield{author}{\bibinfo{person}{Abhishek Das},
  \bibinfo{person}{Th{\'e}ophile Gervet}, \bibinfo{person}{Joshua Romoff},
  \bibinfo{person}{Dhruv Batra}, \bibinfo{person}{Devi Parikh},
  \bibinfo{person}{Mike Rabbat}, {and} \bibinfo{person}{Joelle Pineau}.}
  \bibinfo{year}{2019}\natexlab{}.
\newblock \showarticletitle{Tarmac: Targeted multi-agent communication}. In
  \bibinfo{booktitle}{\emph{International Conference on Machine Learning}}.
  \bibinfo{pages}{1538--1546}.
\newblock


\bibitem[\protect\citeauthoryear{Foerster, Farquhar, Afouras, Nardelli, and
  Whiteson}{Foerster et~al\mbox{.}}{2017}]%
        {foerster2018counterfactual}
\bibfield{author}{\bibinfo{person}{Jakob Foerster}, \bibinfo{person}{Gregory
  Farquhar}, \bibinfo{person}{Triantafyllos Afouras}, \bibinfo{person}{Nantas
  Nardelli}, {and} \bibinfo{person}{Shimon Whiteson}.}
  \bibinfo{year}{2017}\natexlab{}.
\newblock \showarticletitle{Counterfactual multi-agent policy gradients}.
\newblock \bibinfo{journal}{\emph{arXiv preprint arXiv:1705.08926}}
  (\bibinfo{year}{2017}).
\newblock


\bibitem[\protect\citeauthoryear{Ganapathi~Subramanian, Poupart, Taylor, and
  Hegde}{Ganapathi~Subramanian et~al\mbox{.}}{2020}]%
        {ganapathi2020multi}
\bibfield{author}{\bibinfo{person}{Sriram Ganapathi~Subramanian},
  \bibinfo{person}{Pascal Poupart}, \bibinfo{person}{Matthew~E Taylor}, {and}
  \bibinfo{person}{Nidhi Hegde}.} \bibinfo{year}{2020}\natexlab{}.
\newblock \showarticletitle{Multi Type Mean Field Reinforcement Learning}. In
  \bibinfo{booktitle}{\emph{Proceedings of the 19th International Conference on
  Autonomous Agents and MultiAgent Systems}}. \bibinfo{pages}{411--419}.
\newblock


\bibitem[\protect\citeauthoryear{Grau-Moya, Leibfried, and Bou-Ammar}{Grau-Moya
  et~al\mbox{.}}{2018}]%
        {grau2018balancing}
\bibfield{author}{\bibinfo{person}{Jordi Grau-Moya}, \bibinfo{person}{Felix
  Leibfried}, {and} \bibinfo{person}{Haitham Bou-Ammar}.}
  \bibinfo{year}{2018}\natexlab{}.
\newblock \showarticletitle{Balancing two-player stochastic games with soft
  Q-learning}. In \bibinfo{booktitle}{\emph{Proceedings of the 27th
  International Joint Conference on Artificial Intelligence}}.
  \bibinfo{pages}{268--274}.
\newblock


\bibitem[\protect\citeauthoryear{Haarnoja, Tang, Abbeel, and Levine}{Haarnoja
  et~al\mbox{.}}{2017}]%
        {haarnoja2017reinforcement}
\bibfield{author}{\bibinfo{person}{Tuomas Haarnoja}, \bibinfo{person}{Haoran
  Tang}, \bibinfo{person}{Pieter Abbeel}, {and} \bibinfo{person}{Sergey
  Levine}.} \bibinfo{year}{2017}\natexlab{}.
\newblock \showarticletitle{Reinforcement Learning with Deep Energy-Based
  Policies}. In \bibinfo{booktitle}{\emph{International Conference on Machine
  Learning}}. \bibinfo{pages}{1352--1361}.
\newblock


\bibitem[\protect\citeauthoryear{Haarnoja, Zhou, Abbeel, and Levine}{Haarnoja
  et~al\mbox{.}}{2018}]%
        {haarnoja2018soft}
\bibfield{author}{\bibinfo{person}{Tuomas Haarnoja}, \bibinfo{person}{Aurick
  Zhou}, \bibinfo{person}{Pieter Abbeel}, {and} \bibinfo{person}{Sergey
  Levine}.} \bibinfo{year}{2018}\natexlab{}.
\newblock \showarticletitle{Soft Actor-Critic: Off-Policy Maximum Entropy Deep
  Reinforcement Learning with a Stochastic Actor}. In
  \bibinfo{booktitle}{\emph{International Conference on Machine Learning}}.
  \bibinfo{pages}{1861--1870}.
\newblock


\bibitem[\protect\citeauthoryear{Hernandez-Leal, Kartal, and
  Taylor}{Hernandez-Leal et~al\mbox{.}}{2019}]%
        {hernandez2019survey}
\bibfield{author}{\bibinfo{person}{Pablo Hernandez-Leal},
  \bibinfo{person}{Bilal Kartal}, {and} \bibinfo{person}{Matthew~E Taylor}.}
  \bibinfo{year}{2019}\natexlab{}.
\newblock \showarticletitle{A survey and critique of multiagent deep
  reinforcement learning}.
\newblock \bibinfo{journal}{\emph{Autonomous Agents and Multi-Agent Systems}}
  \bibinfo{volume}{6}, \bibinfo{number}{33} (\bibinfo{year}{2019}),
  \bibinfo{pages}{750--797}.
\newblock


\bibitem[\protect\citeauthoryear{Houthooft, Chen, Duan, Schulman, De~Turck, and
  Abbeel}{Houthooft et~al\mbox{.}}{2016}]%
        {houthooft2016vime}
\bibfield{author}{\bibinfo{person}{Rein Houthooft}, \bibinfo{person}{Xi Chen},
  \bibinfo{person}{Yan Duan}, \bibinfo{person}{John Schulman},
  \bibinfo{person}{Filip De~Turck}, {and} \bibinfo{person}{Pieter Abbeel}.}
  \bibinfo{year}{2016}\natexlab{}.
\newblock \showarticletitle{Vime: Variational information maximizing
  exploration}. In \bibinfo{booktitle}{\emph{Advances in Neural Information
  Processing Systems}}. \bibinfo{pages}{1109--1117}.
\newblock


\bibitem[\protect\citeauthoryear{Iqbal, de~Witt, Peng, B{\"o}hmer, Whiteson,
  and Sha}{Iqbal et~al\mbox{.}}{2020}]%
        {iqbal2020ai}
\bibfield{author}{\bibinfo{person}{Shariq Iqbal}, \bibinfo{person}{Christian
  A~Schroeder de Witt}, \bibinfo{person}{Bei Peng}, \bibinfo{person}{Wendelin
  B{\"o}hmer}, \bibinfo{person}{Shimon Whiteson}, {and} \bibinfo{person}{Fei
  Sha}.} \bibinfo{year}{2020}\natexlab{}.
\newblock \showarticletitle{AI-QMIX: Attention and Imagination for Dynamic
  Multi-Agent Reinforcement Learning}.
\newblock \bibinfo{journal}{\emph{arXiv preprint arXiv:2006.04222}}
  (\bibinfo{year}{2020}).
\newblock


\bibitem[\protect\citeauthoryear{Iqbal and Sha}{Iqbal and Sha}{2019}]%
        {iqbal2019actor}
\bibfield{author}{\bibinfo{person}{Shariq Iqbal} {and} \bibinfo{person}{Fei
  Sha}.} \bibinfo{year}{2019}\natexlab{}.
\newblock \showarticletitle{Actor-attention-critic for multi-agent
  reinforcement learning}. In \bibinfo{booktitle}{\emph{International
  Conference on Machine Learning}}. \bibinfo{pages}{2961--2970}.
\newblock


\bibitem[\protect\citeauthoryear{Liu, Zhou, Zhang, Zhuang, Wang, Liu, and
  Yu}{Liu et~al\mbox{.}}{2020}]%
        {liu2019multi}
\bibfield{author}{\bibinfo{person}{Minghuan Liu}, \bibinfo{person}{Ming Zhou},
  \bibinfo{person}{Weinan Zhang}, \bibinfo{person}{Yuzheng Zhuang},
  \bibinfo{person}{Jun Wang}, \bibinfo{person}{Wulong Liu}, {and}
  \bibinfo{person}{Yong Yu}.} \bibinfo{year}{2020}\natexlab{}.
\newblock \showarticletitle{Multi-Agent Interactions Modeling with Correlated
  Policies}.
\newblock \bibinfo{journal}{\emph{arXiv preprint arXiv:2001.03415}}
  (\bibinfo{year}{2020}).
\newblock


\bibitem[\protect\citeauthoryear{Long, Zhou, Gupta, Fang, Wu, and Wang}{Long
  et~al\mbox{.}}{2020}]%
        {long2019evolutionary}
\bibfield{author}{\bibinfo{person}{Qian Long}, \bibinfo{person}{Zihan Zhou},
  \bibinfo{person}{Abhibav Gupta}, \bibinfo{person}{Fei Fang},
  \bibinfo{person}{Yi Wu}, {and} \bibinfo{person}{Xiaolong Wang}.}
  \bibinfo{year}{2020}\natexlab{}.
\newblock \showarticletitle{Evolutionary Population Curriculum for Scaling
  Multi-Agent Reinforcement Learning}.
\newblock \bibinfo{journal}{\emph{arXiv preprint arXiv:2003.10423}}
  (\bibinfo{year}{2020}).
\newblock


\bibitem[\protect\citeauthoryear{Lowe, Wu, Tamar, Harb, Abbeel, and
  Mordatch}{Lowe et~al\mbox{.}}{2017}]%
        {lowe2017multi}
\bibfield{author}{\bibinfo{person}{Ryan Lowe}, \bibinfo{person}{Yi~I Wu},
  \bibinfo{person}{Aviv Tamar}, \bibinfo{person}{Jean Harb},
  \bibinfo{person}{OpenAI~Pieter Abbeel}, {and} \bibinfo{person}{Igor
  Mordatch}.} \bibinfo{year}{2017}\natexlab{}.
\newblock \showarticletitle{Multi-agent actor-critic for mixed
  cooperative-competitive environments}. In \bibinfo{booktitle}{\emph{Advances
  in neural information processing systems}}. \bibinfo{pages}{6379--6390}.
\newblock


\bibitem[\protect\citeauthoryear{Mahajan, Rashid, Samvelyan, and
  Whiteson}{Mahajan et~al\mbox{.}}{2019}]%
        {maven}
\bibfield{author}{\bibinfo{person}{Anuj Mahajan}, \bibinfo{person}{Tabish
  Rashid}, \bibinfo{person}{Mikayel Samvelyan}, {and} \bibinfo{person}{Shimon
  Whiteson}.} \bibinfo{year}{2019}\natexlab{}.
\newblock \showarticletitle{Maven: Multi-agent variational exploration}. In
  \bibinfo{booktitle}{\emph{Advances in Neural Information Processing
  Systems}}. \bibinfo{pages}{7613--7624}.
\newblock


\bibitem[\protect\citeauthoryear{Mao, Zhang, Xiao, and Gong}{Mao
  et~al\mbox{.}}{2019}]%
        {mao2019modelling}
\bibfield{author}{\bibinfo{person}{Hangyu Mao}, \bibinfo{person}{Zhengchao
  Zhang}, \bibinfo{person}{Zhen Xiao}, {and} \bibinfo{person}{Zhibo Gong}.}
  \bibinfo{year}{2019}\natexlab{}.
\newblock \showarticletitle{Modelling the Dynamic Joint Policy of Teammates
  with Attention Multi-agent DDPG}. In \bibinfo{booktitle}{\emph{Proceedings of
  the 18th International Conference on Autonomous Agents and MultiAgent
  Systems}}. \bibinfo{pages}{1108--1116}.
\newblock


\bibitem[\protect\citeauthoryear{Oh, Chockalingam, Lee, et~al\mbox{.}}{Oh
  et~al\mbox{.}}{2016}]%
        {oh2016control}
\bibfield{author}{\bibinfo{person}{Junhyuk Oh}, \bibinfo{person}{Valliappa
  Chockalingam}, \bibinfo{person}{Honglak Lee}, {et~al\mbox{.}}}
  \bibinfo{year}{2016}\natexlab{}.
\newblock \showarticletitle{Control of Memory, Active Perception, and Action in
  Minecraft}. In \bibinfo{booktitle}{\emph{International Conference on Machine
  Learning}}. \bibinfo{pages}{2790--2799}.
\newblock


\bibitem[\protect\citeauthoryear{Oliehoek, Amato, et~al\mbox{.}}{Oliehoek
  et~al\mbox{.}}{2016}]%
        {oliehoek2016concise}
\bibfield{author}{\bibinfo{person}{Frans~A Oliehoek},
  \bibinfo{person}{Christopher Amato}, {et~al\mbox{.}}}
  \bibinfo{year}{2016}\natexlab{}.
\newblock \bibinfo{booktitle}{\emph{A concise introduction to decentralized
  POMDPs}}. Vol.~\bibinfo{volume}{1}.
\newblock \bibinfo{publisher}{Springer}.
\newblock


\bibitem[\protect\citeauthoryear{Oliehoek, Spaan, and Vlassis}{Oliehoek
  et~al\mbox{.}}{2008}]%
        {oliehoek2008optimal}
\bibfield{author}{\bibinfo{person}{Frans~A Oliehoek},
  \bibinfo{person}{Matthijs~TJ Spaan}, {and} \bibinfo{person}{Nikos Vlassis}.}
  \bibinfo{year}{2008}\natexlab{}.
\newblock \showarticletitle{Optimal and Approximate Q-value Functions for
  Decentralized POMDPs}.
\newblock \bibinfo{journal}{\emph{Journal of Artificial Intelligence Research}}
   \bibinfo{volume}{32} (\bibinfo{year}{2008}), \bibinfo{pages}{289--353}.
\newblock


\bibitem[\protect\citeauthoryear{Osband, Blundell, Pritzel, and Van~Roy}{Osband
  et~al\mbox{.}}{2016}]%
        {osband2016deep}
\bibfield{author}{\bibinfo{person}{Ian Osband}, \bibinfo{person}{Charles
  Blundell}, \bibinfo{person}{Alexander Pritzel}, {and}
  \bibinfo{person}{Benjamin Van~Roy}.} \bibinfo{year}{2016}\natexlab{}.
\newblock \showarticletitle{Deep exploration via bootstrapped DQN}. In
  \bibinfo{booktitle}{\emph{Advances in neural information processing
  systems}}. \bibinfo{pages}{4026--4034}.
\newblock


\bibitem[\protect\citeauthoryear{Rashid, Farquhar, Peng, and Whiteson}{Rashid
  et~al\mbox{.}}{2020}]%
        {rashid2020weighted}
\bibfield{author}{\bibinfo{person}{Tabish Rashid}, \bibinfo{person}{Gregory
  Farquhar}, \bibinfo{person}{Bei Peng}, {and} \bibinfo{person}{Shimon
  Whiteson}.} \bibinfo{year}{2020}\natexlab{}.
\newblock \showarticletitle{Weighted QMIX: Expanding Monotonic Value Function
  Factorisation}.
\newblock \bibinfo{journal}{\emph{arXiv preprint arXiv:2006.10800}}
  (\bibinfo{year}{2020}).
\newblock


\bibitem[\protect\citeauthoryear{Rashid, Samvelyan, Schroeder, Farquhar,
  Foerster, and Whiteson}{Rashid et~al\mbox{.}}{2018}]%
        {rashid2018qmix}
\bibfield{author}{\bibinfo{person}{Tabish Rashid}, \bibinfo{person}{Mikayel
  Samvelyan}, \bibinfo{person}{Christian Schroeder}, \bibinfo{person}{Gregory
  Farquhar}, \bibinfo{person}{Jakob Foerster}, {and} \bibinfo{person}{Shimon
  Whiteson}.} \bibinfo{year}{2018}\natexlab{}.
\newblock \showarticletitle{QMIX: Monotonic Value Function Factorisation for
  Deep Multi-Agent Reinforcement Learning}. In
  \bibinfo{booktitle}{\emph{International Conference on Machine Learning}}.
  \bibinfo{pages}{4295--4304}.
\newblock


\bibitem[\protect\citeauthoryear{Son, Kim, Kang, Hostallero, and Yi}{Son
  et~al\mbox{.}}{2019}]%
        {son2019qtran}
\bibfield{author}{\bibinfo{person}{Kyunghwan Son}, \bibinfo{person}{Daewoo
  Kim}, \bibinfo{person}{Wan~Ju Kang}, \bibinfo{person}{David~Earl Hostallero},
  {and} \bibinfo{person}{Yung Yi}.} \bibinfo{year}{2019}\natexlab{}.
\newblock \showarticletitle{QTRAN: Learning to Factorize with Transformation
  for Cooperative Multi-Agent Reinforcement Learning}. In
  \bibinfo{booktitle}{\emph{International Conference on Machine Learning}}.
  \bibinfo{pages}{5887--5896}.
\newblock


\bibitem[\protect\citeauthoryear{Sunehag, Lever, Gruslys, Czarnecki, Zambaldi,
  Jaderberg, Lanctot, Sonnerat, Leibo, Tuyls, et~al\mbox{.}}{Sunehag
  et~al\mbox{.}}{2017}]%
        {sunehag2017value}
\bibfield{author}{\bibinfo{person}{Peter Sunehag}, \bibinfo{person}{Guy Lever},
  \bibinfo{person}{Audrunas Gruslys}, \bibinfo{person}{Wojciech~Marian
  Czarnecki}, \bibinfo{person}{Vinicius Zambaldi}, \bibinfo{person}{Max
  Jaderberg}, \bibinfo{person}{Marc Lanctot}, \bibinfo{person}{Nicolas
  Sonnerat}, \bibinfo{person}{Joel~Z Leibo}, \bibinfo{person}{Karl Tuyls},
  {et~al\mbox{.}}} \bibinfo{year}{2017}\natexlab{}.
\newblock \showarticletitle{Value-decomposition networks for cooperative
  multi-agent learning}.
\newblock \bibinfo{journal}{\emph{arXiv preprint arXiv:1706.05296}}
  (\bibinfo{year}{2017}).
\newblock


\bibitem[\protect\citeauthoryear{Sykora, Ren, and Urtasun}{Sykora
  et~al\mbox{.}}{2020}]%
        {sykora2020multi}
\bibfield{author}{\bibinfo{person}{Quinlan Sykora}, \bibinfo{person}{Mengye
  Ren}, {and} \bibinfo{person}{Raquel Urtasun}.}
  \bibinfo{year}{2020}\natexlab{}.
\newblock \showarticletitle{Multi-agent routing value iteration network}.
\newblock \bibinfo{journal}{\emph{arXiv preprint arXiv:2007.05096}}
  (\bibinfo{year}{2020}).
\newblock


\bibitem[\protect\citeauthoryear{Tang, Houthooft, Foote, Stooke, Chen, Duan,
  Schulman, DeTurck, and Abbeel}{Tang et~al\mbox{.}}{2017}]%
        {tang2017exploration}
\bibfield{author}{\bibinfo{person}{Haoran Tang}, \bibinfo{person}{Rein
  Houthooft}, \bibinfo{person}{Davis Foote}, \bibinfo{person}{Adam Stooke},
  \bibinfo{person}{OpenAI~Xi Chen}, \bibinfo{person}{Yan Duan},
  \bibinfo{person}{John Schulman}, \bibinfo{person}{Filip DeTurck}, {and}
  \bibinfo{person}{Pieter Abbeel}.} \bibinfo{year}{2017}\natexlab{}.
\newblock \showarticletitle{\# exploration: A study of count-based exploration
  for deep reinforcement learning}. In \bibinfo{booktitle}{\emph{Advances in
  neural information processing systems}}. \bibinfo{pages}{2753--2762}.
\newblock


\bibitem[\protect\citeauthoryear{Todorov}{Todorov}{2008}]%
        {todorov2008general}
\bibfield{author}{\bibinfo{person}{Emanuel Todorov}.}
  \bibinfo{year}{2008}\natexlab{}.
\newblock \showarticletitle{General duality between optimal control and
  estimation}. In \bibinfo{booktitle}{\emph{2008 47th IEEE Conference on
  Decision and Control}}. IEEE, \bibinfo{pages}{4286--4292}.
\newblock


\bibitem[\protect\citeauthoryear{Vaswani, Shazeer, Parmar, Uszkoreit, Jones,
  Gomez, Kaiser, and Polosukhin}{Vaswani et~al\mbox{.}}{2017}]%
        {vaswani2017attention}
\bibfield{author}{\bibinfo{person}{Ashish Vaswani}, \bibinfo{person}{Noam
  Shazeer}, \bibinfo{person}{Niki Parmar}, \bibinfo{person}{Jakob Uszkoreit},
  \bibinfo{person}{Llion Jones}, \bibinfo{person}{Aidan~N Gomez},
  \bibinfo{person}{{\L}ukasz Kaiser}, {and} \bibinfo{person}{Illia
  Polosukhin}.} \bibinfo{year}{2017}\natexlab{}.
\newblock \showarticletitle{Attention is all you need}. In
  \bibinfo{booktitle}{\emph{Advances in neural information processing
  systems}}. \bibinfo{pages}{5998--6008}.
\newblock


\bibitem[\protect\citeauthoryear{Walther and Koch}{Walther and Koch}{2007}]%
        {ba2014multiple}
\bibfield{author}{\bibinfo{person}{Dirk~B Walther} {and}
  \bibinfo{person}{Christof Koch}.} \bibinfo{year}{2007}\natexlab{}.
\newblock \showarticletitle{Attention in hierarchical models of object
  recognition}.
\newblock \bibinfo{journal}{\emph{Progress in brain research}}
  \bibinfo{volume}{165} (\bibinfo{year}{2007}), \bibinfo{pages}{57--78}.
\newblock


\bibitem[\protect\citeauthoryear{Wang, Han, Wang, Dong, and Zhang}{Wang
  et~al\mbox{.}}{2020}]%
        {wang2020off}
\bibfield{author}{\bibinfo{person}{Yihan Wang}, \bibinfo{person}{Beining Han},
  \bibinfo{person}{Tonghan Wang}, \bibinfo{person}{Heng Dong}, {and}
  \bibinfo{person}{Chongjie Zhang}.} \bibinfo{year}{2020}\natexlab{}.
\newblock \showarticletitle{Off-Policy Multi-Agent Decomposed Policy
  Gradients}.
\newblock \bibinfo{journal}{\emph{arXiv preprint arXiv:2007.12322}}
  (\bibinfo{year}{2020}).
\newblock


\bibitem[\protect\citeauthoryear{Williams}{Williams}{1992}]%
        {1992Simple}
\bibfield{author}{\bibinfo{person}{Ronald~J. Williams}.}
  \bibinfo{year}{1992}\natexlab{}.
\newblock \showarticletitle{Simple statistical gradient-following algorithms
  for connectionist reinforcement learning}.
\newblock \bibinfo{journal}{\emph{Machine Learning}} \bibinfo{volume}{8},
  \bibinfo{number}{3-4} (\bibinfo{year}{1992}), \bibinfo{pages}{229--256}.
\newblock


\bibitem[\protect\citeauthoryear{Yang, Hao, Chen, Tang, Chen, Hu, Fan, and
  Wei}{Yang et~al\mbox{.}}{2020}]%
        {yang2020q}
\bibfield{author}{\bibinfo{person}{Yaodong Yang}, \bibinfo{person}{Jianye Hao},
  \bibinfo{person}{Guangyong Chen}, \bibinfo{person}{Hongyao Tang},
  \bibinfo{person}{Yingfeng Chen}, \bibinfo{person}{Yujing Hu},
  \bibinfo{person}{Changjie Fan}, {and} \bibinfo{person}{Zhongyu Wei}.}
  \bibinfo{year}{2020}\natexlab{}.
\newblock \showarticletitle{Q-value Path Decomposition for Deep Multiagent
  Reinforcement Learning}.
\newblock \bibinfo{journal}{\emph{arXiv preprint arXiv:2002.03950}}
  (\bibinfo{year}{2020}).
\newblock


\bibitem[\protect\citeauthoryear{Yang, Luo, Li, Zhou, Zhang, and Wang}{Yang
  et~al\mbox{.}}{2018}]%
        {yang2018mean}
\bibfield{author}{\bibinfo{person}{Yaodong Yang}, \bibinfo{person}{Rui Luo},
  \bibinfo{person}{Minne Li}, \bibinfo{person}{Ming Zhou},
  \bibinfo{person}{Weinan Zhang}, {and} \bibinfo{person}{Jun Wang}.}
  \bibinfo{year}{2018}\natexlab{}.
\newblock \showarticletitle{Mean Field Multi-Agent Reinforcement Learning}. In
  \bibinfo{booktitle}{\emph{International Conference on Machine Learning}}.
  \bibinfo{pages}{5571--5580}.
\newblock


\bibitem[\protect\citeauthoryear{Zhang, Yang, Liu, Zhang, and Basar}{Zhang
  et~al\mbox{.}}{2018}]%
        {zhang2018fully}
\bibfield{author}{\bibinfo{person}{Kaiqing Zhang}, \bibinfo{person}{Zhuoran
  Yang}, \bibinfo{person}{Han Liu}, \bibinfo{person}{Tong Zhang}, {and}
  \bibinfo{person}{Tamer Basar}.} \bibinfo{year}{2018}\natexlab{}.
\newblock \showarticletitle{Fully Decentralized Multi-Agent Reinforcement
  Learning with Networked Agents}. In \bibinfo{booktitle}{\emph{International
  Conference on Machine Learning}}. \bibinfo{pages}{5872--5881}.
\newblock


\bibitem[\protect\citeauthoryear{Zhang, Zhang, and Lin}{Zhang
  et~al\mbox{.}}{2019}]%
        {zhang2019efficient}
\bibfield{author}{\bibinfo{person}{Sai~Qian Zhang}, \bibinfo{person}{Qi Zhang},
  {and} \bibinfo{person}{Jieyu Lin}.} \bibinfo{year}{2019}\natexlab{}.
\newblock \showarticletitle{Efficient communication in multi-agent
  reinforcement learning via variance based control}. In
  \bibinfo{booktitle}{\emph{Advances in Neural Information Processing
  Systems}}. \bibinfo{pages}{3235--3244}.
\newblock


\bibitem[\protect\citeauthoryear{Zhang, Yang, and Zha}{Zhang
  et~al\mbox{.}}{2020}]%
        {zhang2020integrating}
\bibfield{author}{\bibinfo{person}{Zhi Zhang}, \bibinfo{person}{Jiachen Yang},
  {and} \bibinfo{person}{Hongyuan Zha}.} \bibinfo{year}{2020}\natexlab{}.
\newblock \showarticletitle{Integrating Independent and Centralized Multi-agent
  Reinforcement Learning for Traffic Signal Network Optimization}. In
  \bibinfo{booktitle}{\emph{Proceedings of the 19th International Conference on
  Autonomous Agents and MultiAgent Systems}}. \bibinfo{pages}{2083--2085}.
\newblock


\bibitem[\protect\citeauthoryear{Ziebart, Maas, Bagnell, and Dey}{Ziebart
  et~al\mbox{.}}{2008}]%
        {ziebart2008maximum}
\bibfield{author}{\bibinfo{person}{Brian~D Ziebart}, \bibinfo{person}{Andrew~L
  Maas}, \bibinfo{person}{J~Andrew Bagnell}, {and} \bibinfo{person}{Anind~K
  Dey}.} \bibinfo{year}{2008}\natexlab{}.
\newblock \showarticletitle{Maximum entropy inverse reinforcement learning}. In
  \bibinfo{booktitle}{\emph{AAAI}}, Vol.~\bibinfo{volume}{8}. Chicago, IL, USA,
  \bibinfo{pages}{1433--1438}.
\newblock


\end{thebibliography}

%%%%%%%%%%%%%%%%%%%%%%%%%%%%%%%%%%%%%%%%%%%%%%%%%%%%%%%%%%%%%%%%%%%%%%%%
\clearpage
\begin{appendices}

\section{QMIXs}\label{QMIXs}
Different from the Equation~\ref{IGM} of QMIX, agents can communicate action-information when calculating individual action-value function: $Q_i(o_i,u_i,u_{-i})$. 
Since the grid task has no state and is the one-step task, the individual action-value function can be simplified: $Q_i(u_i,u_{-i})$.
QMIXs is trained end-to-end to minimise the following loss:
\begin{equation}
    L=\sum_{j=1}^{|b|}\left[\left(r-\sum_{i=1}^{n}\rm{MIX}\left(Q_i(u_i,u_{-i})\right)\right)^2 \right].
\end{equation}
where $b$ is the batch sample from the replay buffer, $n$ is the number of agents and $r$ is the reward. MIX denotes the QMIX neural network.

\section{Experimental Details}\label{Experimental Details}
Hyper-parameters used in experiments and environment specific parameters are listed in Table~\ref{Hyper-parameters} and Table~\ref{Environment Parameters}.
\begin{table}[H] 
    \caption{Hyper-parameters Sheet}
    \label{Hyper-parameters}
    \begin{tabular}{ll}  
    \toprule   
    Hyper-parameter & Value\\
    \midrule   
    $Shared$ & \\  
    \hspace{0.3cm} Policy network learning rate &  $10^{-4}$\\
    \hspace{0.3cm} Value network learning rate & $10^{-3}$\\
    \hspace{0.3cm} Optimizer & Adam\\
    \hspace{0.3cm} Discount factor & 0.95\\
    \hspace{0.3cm} Batch size & 256\\
    \hspace{0.3cm} Replay buffer size & $5\times10^{5}$\\
    \hspace{0.3cm} Parameters update rate & 200\\
    \hline
    $IAC\ \&\ SQ$  & \\
    \hspace{0.3cm} Attention Heads & 4\\
    \hspace{0.3cm} factor in Covariance diagonal matrix & $10^{-6}$\\
    \hspace{0.3cm} $m$ in covariance factor & 1\\
    \bottomrule  
    \end{tabular}
\end{table}

\begin{table}[H]  
    \caption{Environment Parameters}
    \label{Environment Parameters}
    \begin{tabular}{ccc} 
    \toprule   
    Environment & Temperature Parameter & Episode length\\
    \midrule  
    Simple World & 0.02 & 1 \\  
    Classic Control & 0.01 & 200\\  
    Predator-prey &  0.01 & 100 \\
    PO-Predator-prey & 0.01 & 100 \\
    \bottomrule  
    \end{tabular}
\end{table}

For all algorithms, we use fully-connected neural networks as function approximation with 64 hidden units and ReLU as activation function.
The Network Structure of all algorithms is listed in Table~\ref{Network Structure}.
The Collaborative Exploration Net outputs the collaborative exploration matrix $L_{nd\times m}$. We limit the factor in $L_{nd\times m}$ to 1.
We adopt five different random seeds to evaluate all algorithms. Each algorithm is evaluated for 10 rounds and the average reward is taken.

\begin{table}[H]
    \caption{Network Structure}
    \begin{tabular}{lc}
    \toprule   
    Network & Hidden layers \\
    \midrule  
    Individual Policy Net  & 3\\
    Individual Action-Value Net & 3\\
    Collaborative Exploration Net & 2\\
    QMIX Net & 2\\
    $Attention\ Module$ & \\
    \hspace{0.3cm} Encoding Layer & 1\\
    \hspace{0.3cm} Mixing Layer & 2\\
    \hspace{0.3cm} Output Layer & 2\\
    \bottomrule  
    \end{tabular}
    \label{Network Structure}
\end{table}
The above experimental Setting is suitable for Classic Control, Predator-prey, and PO-Predator-prey tasks. For the simple world, we set the individual policy  to one parameter and the hidden units to 10.

\section{Multi-agent Environments}\label{Multi-agent Environments}

\begin{figure}[H]
    \centering
    \includegraphics[width=2.0in]{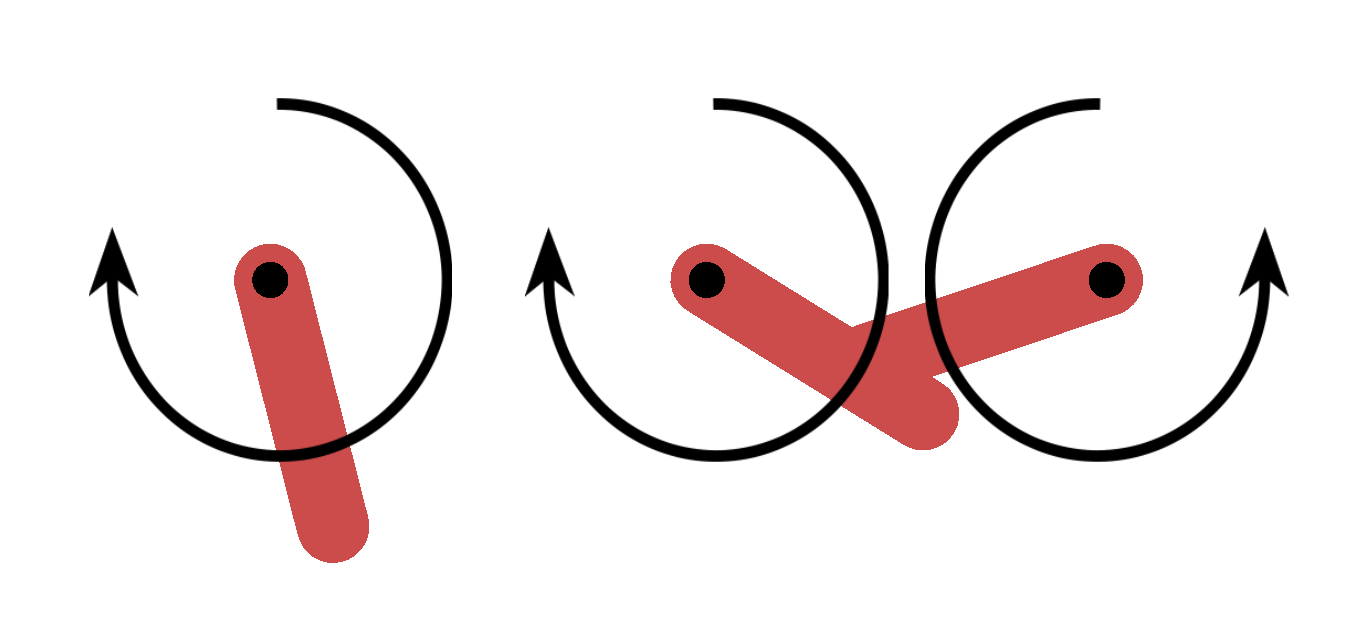}
    \caption{Multi-pendulum task. The number of pendulums can be changed.}
    \label{fig:my_label}
\end{figure}

\begin{figure}[H]
	\centering
	\subfloat[Predator-prey environment.]{
	\includegraphics[width=1.5in]{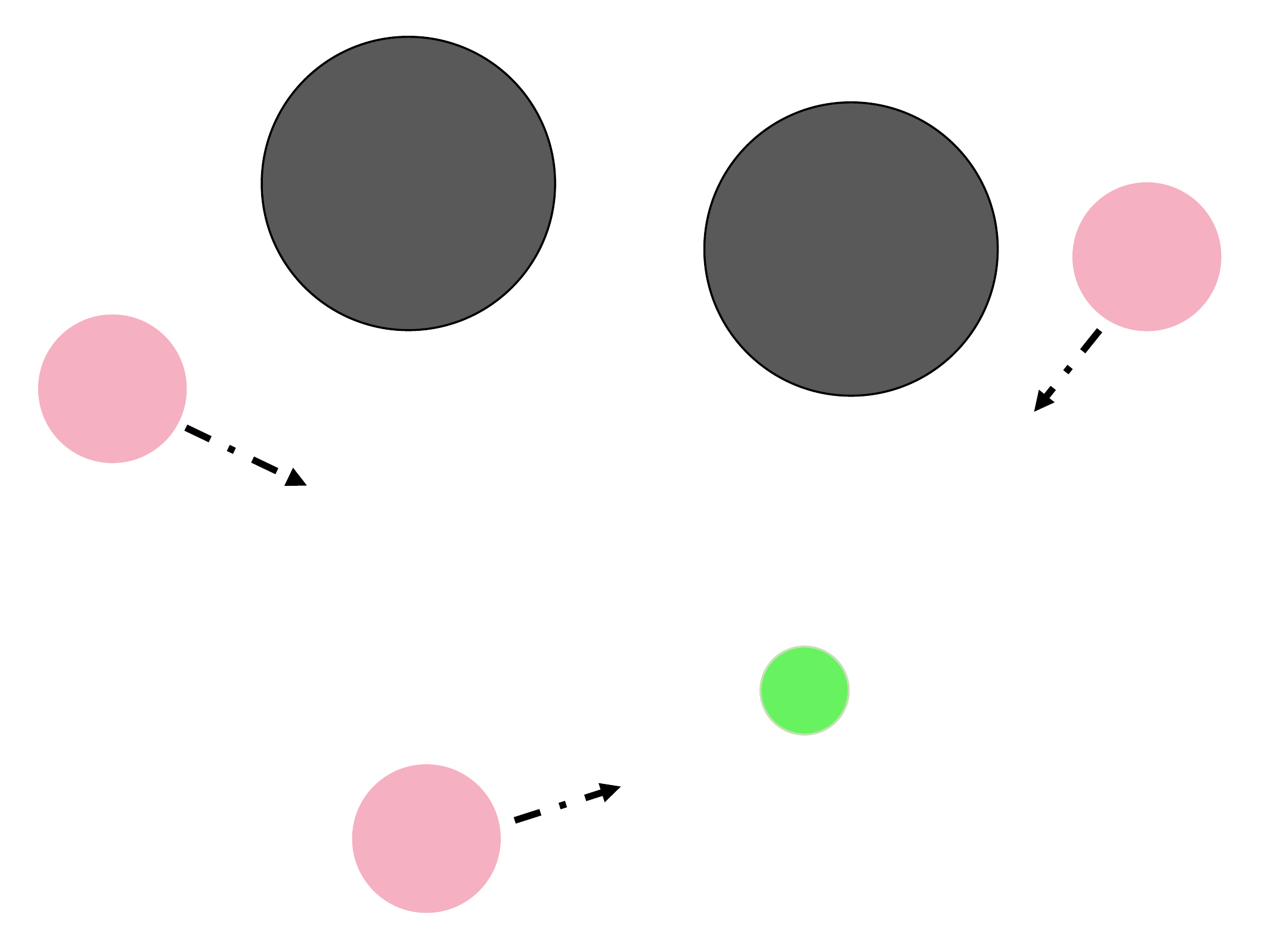}
	}
	\subfloat[PO-Predator-prey environment. 
 	The observation range of agents is marked by a dotted line.]{
	\includegraphics[width=1.5in]{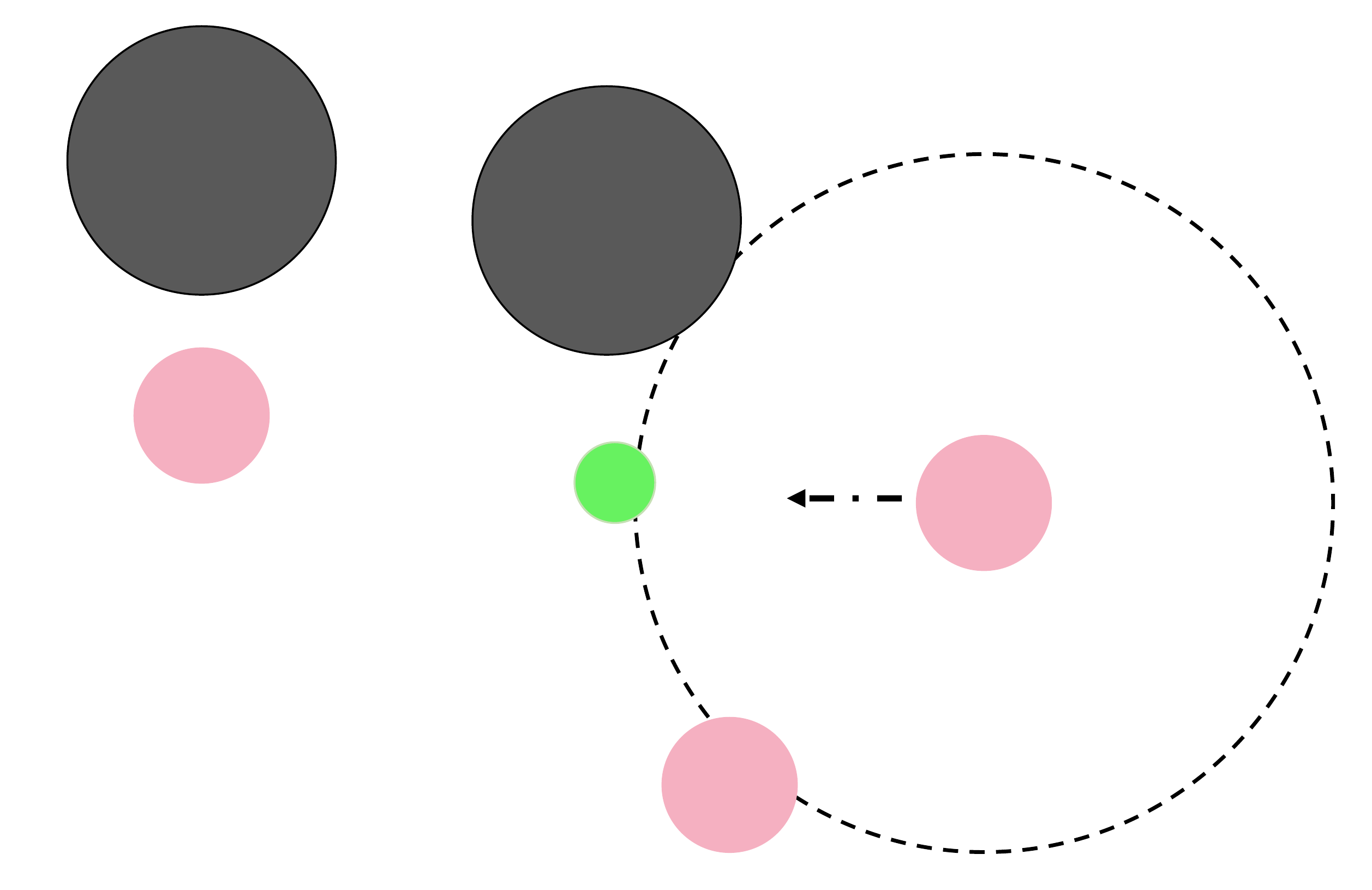}
	}
\caption{Multi-agent particle environment. 
The colors of cooperating agents, adversary agent and landmarks are pink, green and black respectively.
}
\label{predator-prey}
\end{figure}

\end{appendices}
\end{document}